\documentclass[10pt,journal,compsoc]{IEEEtran}
\usepackage{amsmath,amsfonts}
\usepackage{algorithmic}
\usepackage{algorithm}
\usepackage{array}
\usepackage{textcomp}
\usepackage{stfloats}
\usepackage{url}
\usepackage{verbatim}

\usepackage{subcaption}
\usepackage{hyperref}
\usepackage{cite}
\usepackage{multirow}
\usepackage{booktabs}
\usepackage{longtable}
\usepackage{todonotes}
\usepackage{balance}

\usepackage{bm}
\usepackage{cleveref}
\usepackage{standalone}
\usepackage{xspace}
\usepackage[numbers]{natbib}
\usepackage{svg}
\usepackage{graphicx}
\usepackage{pgfplots}
\usepackage{tikz}

\newcommand{\indep}{\perp\!\!\!\perp}

\newcommand\scalemath[2]{\scalebox{#1}{\mbox{\ensuremath{\displaystyle #2}}}}

\begin{document}

\title{Fairness and Bias Mitigation in Computer Vision: A Survey}

\author{
Sepehr Dehdashtian\IEEEauthorrefmark{1},
Ruozhen He\IEEEauthorrefmark{1},
Yi Li,
Guha Balakrishnan,
Nuno Vasconcelos,~\IEEEmembership{Fellow,~IEEE,}\\
Vicente Ordonez,~\IEEEmembership{Member,~IEEE,}
and~Vishnu Naresh Boddeti~\IEEEmembership{Member,~IEEE}%
\IEEEcompsocitemizethanks{\IEEEcompsocthanksitem S. Dehdashtian and V. N. Boddeti are with the Department of Computer Science and Engineering at Michigan State University, East Lansing, Michigan. R. He and V. Ordonez are with the Department of Computer Science at Rice University, Houston, Texas. G. Balakrishnan is with the Department of Electrical and Computer Engineering at Rice University, Houston, Texas. Y. Li and N. Vasconcelos are with the Department of Electrical and Computer Engineering at the University of California, San Diego, California. \protect\\
E-mail: sepehr, visnhu@msu.edu, vicenteor, catherine.he, guha@rice.edu, yil898, nuno@ucsd.edu}
}

\markboth{IEEE Transactions on Pattern Analysis and Machine Intelligence (Under Review)}%
{Dehdashtian \MakeLowercase{\textit{et al.}}: Fairness and Bias Mitigation in Computer Vision: A Survey}

\IEEEtitleabstractindextext{%
\begin{abstract}
Computer vision systems have witnessed rapid progress over the past two decades due to multiple advances in the field. As these systems are increasingly being deployed in high-stakes real-world applications, there is a dire need to ensure that they do not propagate or amplify any discriminatory tendencies in historical or human-curated data or inadvertently learn biases from spurious correlations. This paper presents a comprehensive survey on fairness that summarizes and sheds light on ongoing trends and successes in the context of computer vision. The topics we discuss include 1) The origin and technical definitions of fairness drawn from the wider fair machine learning literature and adjacent disciplines. 2) Work that sought to discover and analyze biases in computer vision systems. 3) A summary of methods proposed to mitigate bias in computer vision systems in recent years. 4) A comprehensive summary of resources and datasets produced by researchers to measure, analyze, and mitigate bias and enhance fairness. 5) Discussion of the field's success, continuing trends in the context of multimodal foundation and generative models, and gaps that still need to be addressed. The presented characterization should help researchers understand the importance of identifying and mitigating bias in computer vision and the state of the field and identify potential directions for future research.

\end{abstract}

}
\maketitle

\begingroup\renewcommand\thefootnote{\IEEEauthorrefmark{1} }
\footnotetext{Equal contribution}
\endgroup

\section{Introduction \label{sec:introduction}}

\IEEEPARstart{T}{he} field of computer vision has gone through several major advances throughout the years. The introduction of machine learning and statistical methods created a wave of interest and progress in visual recognition,~e.g.~\cite{rowley1998neural,viola2001rapid,dalal2005histograms}, which eventually motivated much of the recent advances in deep learning methods using neural networks~\cite{alexnet,resnet,vit} and large-scale datasets~\cite{russakovsky2015imagenet,lin2014microsoft}. The rapid progress in the recognition problem also inspired a search for the right methods and models for a diverse array of other problems, such as U-Nets~\cite{unet} for image segmentation or Latent Diffusion Models~\cite{ldm} for image synthesis. 

Machine learning and statistical methods, however, rely on training datasets and loss functions that can induce, propagate, or magnify statistical biases. Such biases are undesirable when correlated to {\em sensitive protected attributes} such as demographic variables related to people, e.g.~race, gender, age, or ethnicity. Models that learn the inherent correlations or rely on spurious correlations with these attributes can produce disparate outcomes, 
thereby leading to ethical or legal concerns~\cite{howard2019nsf,angwin2022machine}. The goal of {\it fairness and bias mitigation\/}~\cite{mhasawade2021machine,dehdashtian2024utilityfairness} is to prevent or minimize the impact of such biases on model decisions. 

To make computer vision systems widely adopted, accepted, and trusted, it is necessary to avoid societal inequalities and enhance their reliability. This has motivated interest in issues of fairness and biases, intending to develop responsible visual recognition and related systems capable of serving society equitably. From early studies revealing biases in image captioning~\cite{hendricks2018women} or face recognition~\cite{buolamwini2018gender} to recent efforts in mitigating biases in various tasks~\cite{li2023mitigating,yang2023good,dehdashtian2024fairerclip,dehdashtian2024utilityfairness}, there has been a significant body of work in studying fairness and proposing bias mitigation methods for computer vision. In this paper, we survey this literature and related problems solved by machine learning systems trained with large-scale datasets for applications where societal biases are relevant.

The survey first introduces the notation, origins, and definitions of fairness while summarizing the commonalities with fairness studies in the broader machine-learning literature. Then, we briefly discuss prior work on discovering and analyzing bias in computer vision datasets and models. We then present a synthesis of the proposed methodologies and datasets used to study bias and its mitigation. Finally, we discuss current trends in discovering and mitigating bias in multimodal foundation models and open problems in this field. The survey aims to serve as a quick reference and starting point for new research on adapting or designing novel methods to maximize the fairness of emerging computer vision models in a rapidly evolving space. 

What makes the study of fairness in computer vision models distinct from those in other domains, such as tabular data and graphs? The general framework of fairness consists of quantifying the disparate outcomes from a model for groups belonging to different categories of a {\em sensitive protected attribute} and proposing methods to alleviate or mitigate these disparities. For instance, COMPAS~\cite{Angwin2016-lc}, a commonly used tabular dataset to analyze fairness in machine learning, uses race as a sensitive protected attribute, which is included as a categorical variable. In contrast, computer vision datasets usually lack explicit categorical labels for sensitive attributes. Instead, these attributes are implicitly encoded in the combination of input image pixels and task-specific target attributes that are to be inferred by a model. For instance, in the absence of bias mitigation, a computer vision model trained to predict human activities from images, e.g.~{\em cooking vs. not-cooking,} will likely predict the activities at disparate rates for images depicting people of different {\em gender}~\cite{zhao2017men}. The challenge is to disentangle the effect of people's appearance, which is typically correlated with gender, and the activities being performed. Since this is a difficult goal, bias mitigation in computer vision presents unique challenges that are not present in tabular datasets. This justifies a comprehensive survey of computer vision methods, with a brief review of the more general literature on fairness. For a comprehensive survey on fairness in machine learning, we refer the reader to \citet{mehrabi2021survey,pessach2022review,le2022survey,caton2024fairness}. Perhaps more related and complementary to ours is the recent survey by \citet{parraga2023fairness}, which focuses on vision-and-language models. In contrast, our survey provides a more comprehensive summary of the fairness literature related to more traditional computer vision tasks such as image classification, object detection, activity recognition, and face recognition and analysis. 

\begin{figure*}[t]
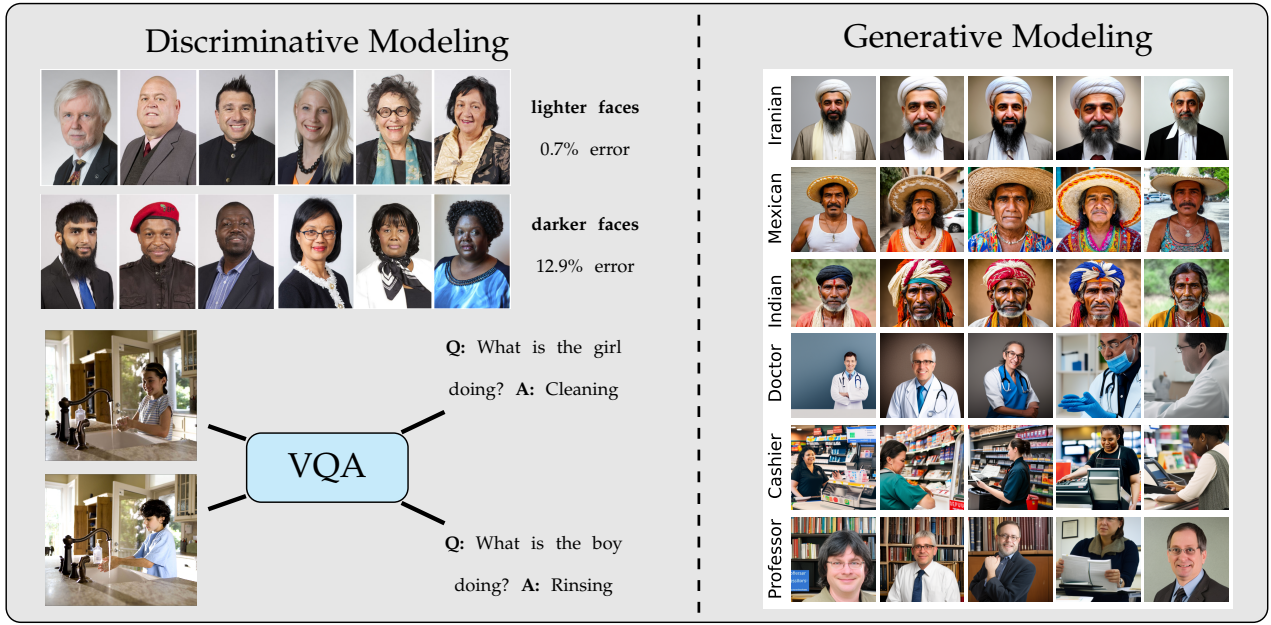

    \centering
    \hspace{-4em}\includestandalone[width=\textwidth]{figs/overview}    
    \caption{Examples of Bias and Unfairness in discriminative and generative computer vision systems. \textbf{Left:} Bias in discriminative modeling shown through face recognition~\cite{buolamwini2018gender} and Situation Recognition~\cite{zhao2017men,yatskar2017commonly} examples. \textbf{Right:} Stereotypical bias in generative modeling with examples from three cultures and three professions~\cite{bianchi2023easily}.
    \label{fig:overview}}
    \vspace{-1.5em}
\end{figure*}
Another challenge in computer vision is the lack of access to explicit labels for {\em sensitive protected attributes}. Commonly, information for demographic variables such as gender, race, or ethnicity is not annotated or provided explicitly by the same individuals depicted in computer vision datasets. Therefore, most annotations on these datasets can only be considered as proxies for the real values based on the perceived judgments of data annotators. Moreover, \citet{scheuerman2024products} argue that tech workers and scientists have also had a significant role in defining categories related to identity for people in computer vision datasets. As a result, demographic markers such as gender have only been studied as binary variables for previous works, and race is often studied as a set of discrete categories. Several works summarized in this survey acknowledge some of these issues, but the overall field should be judged in this context.

Beyond these issues, navigating the challenges of fairness and bias mitigation in computer vision is still a complex endeavor due to the nature of biases, the diversity of datasets and tasks, and trade-offs between model performance and fairness. This survey explores core computer vision tasks and identifies primary challenges associated with achieving fairness and mitigating biases for each task. \Cref{fig:overview} illustrates the type of demographic bias and unfairness prevalent in computer vision systems. \Cref{tab:long,tab:task-datasets} extensively summarize task-specific debiasing methods developed in the computer vision literature and the associated datasets employed for studying bias and fairness, respectively. A detailed overview of common methods for bias mitigation and a comprehensive discussion of the datasets categorized by bias attributes and tasks can be found in \Cref{sec:method} and \Cref{sec:datasets}, respectively.

\begin{table*}%
    \renewcommand{\arraystretch}{1.1} %
    
    \caption{\textbf{Methods}: Bias analysis and mitigation techniques by task and protected attributes. While task-specific bias mitigation methods have been proposed bias mitigation for generic visual representation learning received a lot of attention.\label{tab:attr-table}} \label{tab:long}
    \begin{tabular}{p{3cm}p{3cm}p{10.4cm}}
    \toprule
    Task & Attribute & References \\
    \midrule

    \multirow{20}{*}{Representation Learning} & 
    Gender &
    \citet{park2023training},
    \citet{li2023partition},
    \citet{dehdashtian2024fairerclip},
    \citet{dehdashtian2024utilityfairness},
    \citet{sadeghi2022on},
    \citet{qraitem2023bias},
    \citet{jang2023difficulty},
    \citet{wang2023overwriting},
    \citet{zhang2023learning},
    \citet{tang2023fair}, 
    \citet{meister2023gender},
    \citet{ranjit2023variation},
    \citet{jeon2022conservative},
    \citet{park2022fair},
    \citet{wang2022fairness},
    \citet{seo2022unsupervised},
    \citet{chai2022self},
    \citet{zhu2021learning},
    \citet{tartaglione2021end},
    \citet{ramaswamy2021fair},
    \citet{kim2021biaswap},
    \citet{wang2021directional},
    \citet{wang2020towards},
    \citet{wang2019balanced},
    \citet{seth2023dear},
    \citet{hall2024visogender},
    \citet{chuang2023debiasing},
    \citet{van2016stereotyping} \\
    
    & Color &
    \citet{park2023training},
    \citet{jang2023difficulty},
    \citet{zhang2023learning},
    \citet{park2022fair},
    \citet{seo2022unsupervised},
    \citet{zhu2021learning},
    \citet{wang2021causal},
    \citet{jung2021fair},
    \citet{tartaglione2021end},
    \citet{kim2021biaswap},
    \citet{wang2020towards},
    \citet{li2019repair} \\
    & Corruption &
    \citet{park2023training}, \citet{zhang2023learning} \\
    & Age &
    \citet{li2023partition}, 
    \citet{dehdashtian2024utilityfairness},
    \citet{sadeghi2022on},
    \citet{qraitem2023bias},
    \citet{park2022fair},
    \citet{zhu2021learning},
    \citet{tartaglione2021end},
    \citet{seth2023dear},
    \citet{chuang2023debiasing} \\
    
    & Race &
    \citet{qraitem2023bias},
    \citet{dehdashtian2024fairerclip},
    \citet{dehdashtian2024utilityfairness},
    \citet{sadeghi2022on},
    \citet{wang2023overwriting},
    \citet{park2022fair},
    \citet{chai2022self},
    \citet{zhu2021learning},
    \citet{seth2023dear},
    \citet{chuang2023debiasing},
    \citet{van2016stereotyping} \\
    
    & Geography &
    \citet{wang2023overwriting},
    \citet{shankar2017no},
    \citet{wang2022revise} \\
    
    & Context &
    \citet{zhang2023learning}, \citet{wang2021causal}, \citet{chuang2023debiasing}, \citet{wang2022revise} \\
    & Scene &\citet{mo2021object} \\
    & Skin Tone &\citet{schumann2024consensus} \\
    & Texture &\citet{wang2021causal}, \citet{kim2021biaswap} \\
    & Action &\citet{li2019repair} \\
    \midrule

    \multirow{2}{*}{Analysis} & Social & \citet{sirotkin2022study}, \citet{birhane2024dark}, \citet{brinkmann2023multidimensional} \\
      & Gender & 
      \citet{meister2023gender}, 
      \citet{birhane2024dark},
      \citet{iofinova2023bias},
      \citet{guilbeault2024online} \\
    \midrule

    \multirow{10}{*}{Classification} & 
    Gender & 
    \citet{kim2019learning},
    \citet{zietlow2022leveling},
    \citet{bendekgey2021scalable},
    \citet{lee2022viscuit},
    \citet{jung2022learning},
    \citet{zhang2022fairness},
    \citet{roy2019mitigating},
    \citet{sadeghi2019global},
    \citet{dehdashtian2024fairerclip},
    \citet{sadeghi2022on},
    \citet{dehdashtian2024utilityfairness},
    \citet{gustafson2023facet} \\
     & 
    Age & 
    \citet{kim2019learning}, \citet{sadeghi2019global}, \citet{dehdashtian2024fairerclip}, \citet{sadeghi2022on}, \citet{dehdashtian2024utilityfairness}, \citet{gustafson2023facet} \\
    & 
    Race &  \citet{lee2022viscuit}, \citet{jung2022learning}, \citet{dehdashtian2024fairerclip} \\
    & 
    Illumination &  \citet{roy2019mitigating}, \citet{sadeghi2019global} \\
    & Hair Color &  \citet{dehdashtian2024utilityfairness}\\
    & Skin Tone &  \citet{gustafson2023facet} \\
    & Other & \citet{singh2020don}, \citet{kim2019learning}, \citet{chiu2023better}, \citet{jia2022visual}, \citet{li2021discover} \\
    \midrule

        \multirow{2}{*}{Action Recognition} 
    & Scene & \citet{choi2019can}, \citet{zhai2023soar}, \citet{li2023mitigating} \\
     & Contextual & \citet{choi2019can} \\
    \midrule
    
    \multirow{12}{*}{Face Recognition} & 
    Gender & 
    \citet{buolamwini2018gender},
    \citet{vera2019facegenderid}
    \citet{quadrianto2019discovering},
    \citet{domnich2021responsible},
    \citet{dhar2021pass},
    \citet{gong2021mitigating},
    \citet{ma2023invariant},
    \citet{liang2023benchmarking},
    \citet{dooley2024rethinking},
    \citet{chen2021understanding},
    \citet{chouldechova2022unsupervised}, 
    \citet{terhorst2021comprehensive}, 
    \mbox{\citet{shankar2017no}}, 
    \citet{zietlow2022leveling}, %
    \citet{georgopoulos2021mitigating}, \citet{li2022cat},  \citet{gong2020jointly} \\
    
     & Race & 
     \citet{buolamwini2018gender},
     \citet{wang2020mitigating},
     \citet{gong2021mitigating},
     \citet{ma2023invariant},
     \citet{liang2023benchmarking},
     \citet{dooley2024rethinking}, 
     \citet{chouldechova2022unsupervised},
     \citet{terhorst2021comprehensive}, 
     \citet{shankar2017no}, 
     \citet{georgopoulos2021mitigating}, 
     \citet{gong2020jointly}  \\
     & Data Imbalance & \citet{liu2019fair}, \cite{terhorst2021comprehensive}  \\
     & Skin Tone & \citet{balakrishnan2021towards}, \citet{dhar2021pass},  \citet{terhorst2021comprehensive}, \citet{georgopoulos2021mitigating}  \\
     & Age, Hair \& Facial Hair & \citet{balakrishnan2021towards},  \citet{terhorst2021comprehensive}, \citet{shankar2017no}, \citet{georgopoulos2021mitigating}, \citet{gong2020jointly}  \\
     & Other & \citet{terhorst2021comprehensive} \\
     \midrule

    \multirow{5}{*}{Generative Models} & 
    Race & \citet{maluleke2022studying}, \citet{tan2020improving}, \citet{wu2022generative} \\
    & Data Imbalance & \citet{yu2020inclusive}, \citet{zhao2018bias} \\
    & Gender & \citet{xu2018fairgan}, \citet{tan2020improving}, \citet{karakas2022fairstyle}, \citet{choi2020fair}, \citet{wu2022generative} \\
    & Age & \citet{tan2020improving}, \citet{karakas2022fairstyle} \\
     & Other & \citet{jalal2021fairness}, \citet{wu2022generative}, \citet{choi2020fair}, \citet{kenfack2022repfair}, \citet{karakas2022fairstyle}, \citet{tan2020improving} \\
    \midrule
    
    \multirow{2}{*}{Object Detection} & 
    Income & \citet{sudhakar2023icon} \\
    & Skin Tone & \citet{wilson2019predictive} \\
    \midrule

    \multirow{2}{*}{Other} 
    & \multirow{2}{*}{-} & 
    \citet{kong2024mitigating},
    \citet{yenamandra2023facts},
    \citet{qiu2023cafeboost},
    \citet{shankar2017no},
    \citet{chu2021learning},
    \citet{garcia2023uncurated},
    \citet{biswas2023probabilistic},
    \citet{tang2020unbiased} \\
    \bottomrule
    \end{tabular}
\end{table*}

\begin{table*}[!ht]
  \centering
  \caption{\textbf{Datasets:} Tasks, datasets, and sensitive protected attributes studied for bias quantification and mitigation. Next to each dataset, we reference either the original paper or the paper that adapted it specifically for bias analysis. %
  ~\label{tab:task-datasets}}
    \begin{tabular}{p{2.4cm}p{3.4cm}p{10cm}}
      \toprule
      Task & Protected Attribute & Datasets \\
      \midrule

      Basic Image Bias$\quad$ Analysis & 
      Social~Context$^S$, Gender$^G$, Age$^A$, Background$^B$&
      Social Context~\cite{li2017dual}\cite{sirotkin2022study}$^S$, 
      MSCOCO~\cite{lin2014microsoft}\cite{zhao2017men}\cite{meister2023gender}$^G$, 
      OpenImages~\cite{krasin2017openimages}~\cite{meister2023gender}$^G$$^A$, 
      CelebA~\cite{liu2015deep}\cite{iofinova2023bias}$^G$, 
      IAT~\cite{schimmack2021implicit}\cite{guilbeault2024online}$^G$, 
      Waterbird~\cite{sagawa2019distributionally}\cite{yenamandra2023facts}$^B$, 
      NICO++~\cite{zhang2023nico++}\cite{yenamandra2023facts}$^B$ \\
      \midrule

      Representation Learning & 
      Geography$^P$, Gender$^G$, Color$^L$, Background$^B$, Age$^A$, Ethnicity$^E$, Context$^C$, Texture$^X$, Other$^O$&
      ImageNet~\cite{russakovsky2015imagenet}~\cite{shankar2017no}$^P$, 
      Open Images~\cite{krasin2017openimages}~\cite{shankar2017no}$^P$, 
      MSCOCO~\cite{lin2014microsoft}~\cite{zhao2017men}~\cite{wang2019balanced}$^G$, 
      CIFAR-10S~\cite{krizhevsky2009learning}~\cite{wang2020towards}$^L$, 
      Corrupted CIFAR10~\cite{krizhevsky2009learning}~\cite{hendrycks2018benchmarking}~\cite{kim2021biaswap}$^O$, 
      BAR~\cite{nam2020learning}~\cite{kim2021biaswap}$^B$, 
      bFFHQ~\cite{karras2019style}~\cite{kim2021biaswap}$^G$, 
      IMDB Face~\cite{rothe2018deep}~\cite{kim2019learning}~\cite{tartaglione2021end}$^G$$^A$, 
      CelebA~\cite{liu2015deep}~\cite{nam2020learning}~\cite{tartaglione2021end}$^G$, 
      UTKFace~\cite{zhang2017age}~\cite{quadrianto2019discovering}$^E$, 
      NICO~\cite{he2021towards}~\cite{wang2021causal}$^C$, 
      ImageNet-A~\cite{hendrycks2021natural,wang2021causal}$^L$$^X$, 
      mPower~\cite{kim2019learning}
      ~\cite{zhu2021learning}$^A$, 
      Adult~\cite{yurochkin2020training}
      ~\cite{zhu2021learning}
      $^G$$^R$$^O$, 
      LFW~\cite{LFWTech}
      ~\cite{wang2022fairness}$^G$, 
      DollarStreet~\cite{rojas2022dollar}
      ~\cite{wang2023overwriting}$^P$, 
      GeoDE~\cite{ramaswamy2024geode}
      ~\cite{wang2023overwriting}$^P$, 
      MST%
      ~\cite{schumann2024consensus}$^T$, 
      PATA~\cite{seth2023dear}$^C$, 
      VisoGender~\cite{hall2024visogender}
      $^G$ \\
      \midrule
      
      Image $\quad\quad\quad\quad\quad$ Classification & 
      Color$^L$, Age$^A$, Gender$^G$, Context$^C$, Ethnicity$^E$, Background$^B$, Social Context$^S$, Other$^O$ &
      Colored MNIST~\cite{arjovsky2019invariant}~\cite{kim2019learning}$^L$, 
      Dogs and Cats~\cite{dogs-vs-cats}~\cite{kim2019learning}$^L$, 
      IMDB Face~\cite{wang2018devil}~\cite{kim2019learning}$^A$$^G$, 
      MSCOCO-Stuff~\cite{caesar2018cvpr}~\cite{singh2020don}$^C$, 
      UnRel~\cite{Peyre17}~\cite{singh2020don}$^C$, 
      Deep Fashion~\cite{liu2016deepfashion}~\cite{singh2020don}$^C$, 
      AwA~\cite{xian2018zero}~\cite{singh2020don}$^C$, 
      CelebA~\cite{liu2015deep}~\cite{li2021discover}~$^G$$^E$$^A$$^O$, 
      Faces of the World~\cite{escalera2017chalearn}~\cite{bendekgey2021scalable}$^G$, 
      UTKFace~\cite{zhang2017age}\cite{jung2022learning}$^E$$^G$, 
      COMPAS~\cite{angwin2020there}~\cite{jung2022learning}$^E$$^G$, 
      CIFAR-10-B~\cite{krizhevsky2009learning}\cite{chiu2023better}\cite{chiu2023better}$^B$, 
      CIFAR-100-B~\cite{krizhevsky2009learning}\cite{chiu2023better}\cite{chiu2023better}$^B$, 
      Extended Yale B~\cite{Extended-Yale-B}~\cite{roy2019mitigating}$^G$$^L$, 
      Waterbird~\cite{sagawa2019distributionally}~\cite{dehdashtian2024fairerclip}$^B$, 
      CFD~\cite{ma2015chicago}~\cite{dehdashtian2024fairerclip}$^G$, 
      \mbox{FairFace}~\cite{karkkainen2019fairface}~\cite{dehdashtian2024fairerclip}$^S$ \\
      \midrule
      
      Action Recognition & 
      Background$^B$ &
      UCF-101~\cite{soomro2012ucf101}~\cite{choi2019can}$^B$, 
      HMDB-51~\cite{HMDB}~\cite{choi2019can}$^B$, 
      Diving48~\cite{li2018resound}~\cite{choi2019can}$^B$, 
      THUMOS-14~\cite{THUMOS14}~\cite{choi2019can}$^B$, 
      JHMDB~\cite{Jhuang:ICCV:2013}~\cite{choi2019can}$^B$, 
      MiTv2~\cite{monfort2019moments}~\cite{zhai2023soar}$^B$, 
      SCUBA~\cite{li2023mitigating}$^B$, 
      SCUFO~\cite{li2023mitigating}$^B$ \\
      \midrule
      
      Face Recognition$\quad$ and Analysis & 
      Gender$^G$, Race$^R$, Age$^A$, Skin Tone$^T$, Other$^O$ &
      PPB~\cite{buolamwini2018gender}\cite{balakrishnan2021towards}$^G$$^R$, 
      IJB-A~\cite{klare2015pushing}\cite{buolamwini2018gender}$^G$$^R$, 
      Adience~\cite{adience}\cite{buolamwini2018gender}$^G$$^R$, 
      DiF~\cite{merler2019diversity}\cite{quadrianto2019discovering}$^G$, 
      Adult~\cite{yurochkin2020training}\cite{zhu2021learning}$^G$, 
      LFW\cite{LFWTech}\cite{liu2019fair}$^O$, 
      YTF\cite{wolf2011face}\cite{liu2019fair}$^O$, 
      MegaFace\cite{nech2017level}\cite{liu2019fair}$^O$,
      Transect\cite{balakrishnan2021towards}$^G$$^T$$^A$$^O$,
      IJB-C\cite{IJB-C}$^G$$^T$,
      RFW~\cite{wang2019racial}$^G$$^R$, 
      MFR~\cite{deng2021masked}\cite{ma2023invariant}$^R$, 
      IJB-B~\cite{whitelam2017iarpa}\cite{ma2023invariant}$^G$,
      DigiFace-1M~\cite{bae2023digiface1m}\cite{ma2023invariant}$^O$, 
      VGGFace2~\cite{cao2018vggface2}\cite{dooley2024rethinking}$^G$$^R$, 
      KDEF~\cite{lundqvist1998karolinska}\cite{chen2021understanding}$^G$, 
      CFD~\cite{ma2015chicago}\cite{dehdashtian2024fairerclip}$^G$$^R$, 
      ExpW~\cite{zhang2015learning}\cite{chen2021understanding}$^G$, 
      RAF-DB~\cite{li2017reliable}\cite{chen2021understanding}$^G$, 
      AffectNet~\cite{mollahosseini2017affectnet}\cite{chen2021understanding}$^G$, 
      CausalFaces~\cite{liang2023benchmarking}$^G$$^R$, 
      MORPH~\cite{ricanek2006morph}\cite{georgopoulos2021mitigating}$^G$$^R$$^A$, 
      MAAD-Face (47 attributes)~\cite{terhorst2021maad}\cite{terhorst2021comprehensive}, 
      FairFace~\cite{karkkainen2019fairface}$^G$$^R$$^A$, 
      CelebA~\cite{liu2015deep, zietlow2022leveling}\cite{quadrianto2019discovering}$^G$, 
      CACD~\cite{chen2014cross}\cite{georgopoulos2021mitigating}$^G$$^A$, 
      KANFace~\cite{georgopoulos2020investigating}\cite{georgopoulos2021mitigating}$^G$$^A$, 
      UTKFace~\cite{zhang2017age}\cite{jung2022learning}$^G$$^A$$^T$,
      MS1MV2~\cite{deng2019arcface,guo2016ms}\cite{meng2021magface}$^O$, 
      MS-Celeb-1M~\cite{gong2020jointly, deng2019arcface}$^G$$^A$$^R$, 
      CFP-FP\cite{sengupta2016frontal}\cite{meng2021magface} %
      \\
      \midrule

      Image Retrieval & 
      Gender$^G$ &
      Occupation 1~\cite{kay2015unequal}\cite{kong2024mitigating}$^G$, 
      Occupation 2~\cite{celis2020implicit}\cite{kong2024mitigating}$^G$, 
      MSCOCO~\cite{lin2014microsoft}\cite{kong2024mitigating}$^G$, 
      Flickr30k~\cite{plummer2015flickr30k}\cite{kong2024mitigating}$^G$ \\
      \midrule
      
      Object Detection & 
      Skin Tone$^T$, Income$^I$&
      BDD100K~\cite{BDD100K}\cite{wilson2019predictive}$^T$$^I$ \\
      \midrule
      
      Person$\quad\quad\quad\quad$ Reidentification & 
      Clothing$^H$&
      PRCC-ReID~\cite{yang2019person}\cite{yang2023good}$^H$, 
      LTCC-ReID~\cite{qian2020long}\cite{yang2023good}$^H$ \\
      \midrule

      Image Captioning & 
      Gender$^G$, Race$^R$ &
      MSCOCO-Bias~\cite{hendricks2018women}$^G$, 
      MSCOCO-Balanced~\cite{hendricks2018women}$^G$, 
      MSCOCO~\cite{lin2014microsoft}\cite{bhargava2019exposing}$^G$$^R$ \\
      \midrule

      Image Question $\quad$ Answering & 
      Language$^N$, Context$^C$, Gender$^G$, Race$^R$&
      VQA~\cite{VQA}\cite{manjunatha2019explicit}$^N$, 
      MSCOCO~\cite{lin2014microsoft}\cite{manjunatha2019explicit}$^N$, 
      VQA-CP-v2~\cite{agrawal2018don}\cite{kv2020reducing}$^N$, 
      VQA-v2~\cite{goyal2017making}\cite{kv2020reducing}$^N$$^C$, 
      VQA-Gender~\cite{park2020fair}\cite{park2020fair}$^G$, 
      VQA-introspect~\cite{selvaraju2020squinting}$^C$, 
      IV-VQA~\cite{agarwal2020towards}$^C$, 
      CV-VQA~\cite{agarwal2020towards}$^C$, 
      VQA-CP~\cite{agrawal2018don}$^L$, 
      GQA-OOD~\cite{kervadec2021roses}$^L$, 
      VQA-CE~\cite{dancette2021beyond}$^L$, 
      Visual7W~\cite{zhu2016visual7w}$^R$$^G$, 
      OK-VQA~\cite{marino2019ok}$^G$ \\

      \midrule

      Scene Graph $\quad\quad$ Generation & 
      Composition$^M$&
      VG~\cite{krishna2017visual}\cite{hirota2022gender}$^M$, 
      MSCOCO~\cite{lin2014microsoft}\cite{tang2020unbiased}$^M$ \\
      \midrule
      
      Text-to-Image $\quad\quad$ Synthesis& 
      Gender$^G$, Skin Tone$^T$&
      CelebA~\cite{liu2015deep}\cite{zhang2023iti}$^G$, 
      FAIR~\cite{feng2022towards}\cite{zhang2023iti}$^T$, 
      FairFace~\cite{karkkainen2019fairface}\cite{maluleke2022studying}$^G$$^R$ \\

      \bottomrule
    \end{tabular}
\end{table*}

\section{Origins and Definitions of Unfairness}
We start with a note on terminology. The term {\em bias} has been overloaded in the context of the study of fairness. A statistical bias simply refers to the degree to which a certain methodology provides a skewed representation of a true phenomenon. For instance, opinion surveys conducted only through the workplace overlook unemployed people and are thus not representative of the sentiment of the general population. In computer vision, biases can manifest in multiple ways. For example, Torralba~and~Efros~\cite{torralba2011unbiased} studied bias in early visual recognition models, showing that training on a particular dataset did not generalize well to others. Classifiers trained on one dataset were skewed to produce satisfactory results only for images resembling those in that dataset. In this case, the bias is w.r.t. the {\em dataset} variable. Ideally, a classifier trained on a combination of several datasets should perform well across test splits for all datasets. In the context of action recognition, \citet{li2018resound} showed that datasets frequently have clues, such as objects, backgrounds, etc., that enable good recognition performance by video representations that only account for a single or a few video frames. In this case, the bias is w.r.t. the {\em representation} variable. Various datasets~\cite{li2018resound,somethingsomething} have since been introduced to combat this problem by requiring the classification of fine-grained actions, distinguishable only by long-term motion patterns. We will refer to methods proposed to correct biases as {\em mitigation} techniques, which are also often referred to as {\em debiasing} techniques.

References to fairness and mitigating biases in machine learning models are often used interchangeably when bias mitigation targets a {\em sensitive protected attribute}. Typical examples of this type of attribute in computer vision include sensitive demographic variables such as the gender, race/ethnicity, age, and skin tone of people depicted in images. For instance, the work of \citet{buolamwini2018gender} showed disparities in the success rate of a gender classifier depending on the skin tone of the depicted individuals.  However, depending on the context, other variables, such as geographical location, could be considered sensitive protected attributes. For instance, \citet{shankar2017no} uses geo-location as a protected attribute to study disparities in the performance of visual recognition models for images obtained from different parts of the world. Our survey aims to cover bias analysis, and mitigation works that deal with sensitive protected attributes to improve the fairness of computer vision model predictions. However, we also consider work that uses synthetic or simulated protected attributes introduced to study fairness. For instance, \citet{wang2020towards} proposes a variant of the CIFAR-10 dataset where a percentage of images were converted to grayscale and uses {\em coloring} as a protected attribute. Similarly, we also consider work that proposes bias mitigation techniques where the protected attributes are contextual cues from language or image backgrounds that should be irrelevant to the intended task. For instance, \citet{choi2019can} uses the background scene of a video as a protected attribute for the action recognition task. A well-behaved action recognition model should detect an action regardless of the background scene in which it takes place. We also acknowledge that there is significant work outside the scope of this survey that might share a similar methodology but whose main goal is to improve privacy, transparency, or accountability. 

The rest of this section discusses two important aspects of fairness. In Section~\ref{sec:biasorigins}, we discuss factors that contribute to biases in current computer vision models. In~\Cref{sec:fairnessdefinitions}, we discuss criteria used in the literature to define fairness across sensitive protected attributes.

\subsection{Bias Origins}
\label{sec:biasorigins}
It is well documented that multiple machine learning and computer vision models have exhibited biases w.r.t. {\em sensitive protected attributes} in various contexts and applications~e.g., gender or skin tone~\cite{birhane2024dark, wang2021gender, birhane2023hate, birhane2023into, dehdashtian2024utilityfairness} and even non-demographic attributes~e.g., image background or illumination~\cite{du2022learning, zhang2022contrastive}. These biases are a manifestation of both \emph{social} and \emph{machine learning} biases, with the former largely arising from the training data on which the models are trained. %

From a \emph{social perspective}, the world is frequently biased; for example, expensive cars are more common in affluent than poor neighborhoods. These biases can be amplified by the publication of data on the internet, where most of the large public datasets are collected, e.g., expensive car manufacturers or owners tend to post images of the cars against landmarks or beautiful scenery. As a result, biases found in trained models are largely inherited from the data used to train them, which studies have shown to exhibit similar biases~\cite{misra2016seeing,zhao2017men,wang2019balanced,de2019does}. Datasets can also amplify biases due to data collection practices, e.g., data collection predominantly in some countries or continents~\cite{lyons1998coding,lyons2021excavating,singh2020indian}. Ultimately, the biases in the data are either inherited from human biases as reflected on the Internet or the methodology used to collect and annotate it.

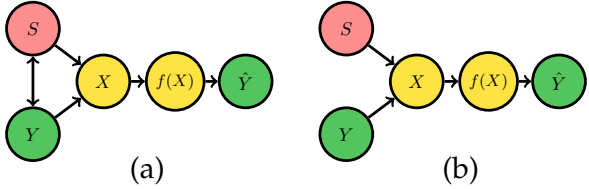
\begin{figure}[t]
    \centering
    \definecolor{color2}{HTML}{f7f8fa} %
\definecolor{color1}{HTML}{DEDEFF} %
\definecolor{color3}{HTML}{FFE342} %
\definecolor{color4}{HTML}{FF9933} %
\definecolor{color5}{HTML}{54c45e} %
\definecolor{color6}{HTML}{ff8f8f} %
\definecolor{color7}{HTML}{CFD3D8} %

\begin{tikzpicture}[node distance=4e]
     \def\lw {1.5}

    \node[draw=black, fill=color3, align=center, anchor=center, circle, line width=\lw pt, minimum width=3em] (x) {$X$};
    \node[draw=black, fill=color3, align=center, anchor=center, circle, line width=\lw pt, minimum width=3em] (f) at ([xshift=2.5em, yshift=0em]x.east) {\small{$f(X)$}};
    \node[draw=black, fill=color6, align=center, anchor=center, circle, line width=\lw pt, minimum width=3em] (s) at ([xshift=-4em, yshift=3em]x) {$S$};
    \node[draw=black, fill=color5, align=center, anchor=center, circle, line width=\lw pt, minimum width=3em] (y) at ([xshift=-4em, yshift=-3em]x) {$Y$};
    \node[draw=black, fill=color5, align=center, anchor=center, circle, line width=\lw pt, minimum width=3em] (yhat) at ([xshift=4em, yshift=0em]f) {$\hat{Y}$};

    \draw[->, line width=\lw pt] (s) -- (x);
    \draw[->, line width=\lw pt] (y) -- (x);
    \draw[<->, line width=\lw pt] (s) -- (y);
    \draw[->, line width=\lw pt] (x) -- (f);
    \draw[->, line width=\lw pt] (f) -- (yhat);

    \node[draw=black, fill=color3, align=center, anchor=center, circle, line width=\lw pt, minimum width=3em] (x1) at ([xshift=12em]f.east) {$X$};
    \node[draw=black, fill=color3, align=center, anchor=center, circle, line width=\lw pt, minimum width=3em] (f1) at ([xshift=2.5em, yshift=0em]x1.east) {\small{$f(X)$}};
    \node[draw=black, fill=color6, align=center, anchor=center, circle, line width=\lw pt, minimum width=3em] (s1) at ([xshift=-4em, yshift=3em]x1) {$S$};
    \node[draw=black, fill=color5, align=center, anchor=center, circle, line width=\lw pt, minimum width=3em] (y1) at ([xshift=-4em, yshift=-3em]x1) {$Y$};
    \node[draw=black, fill=color5, align=center, anchor=center, circle, line width=\lw pt, minimum width=3em] (yhat1) at ([xshift=4em, yshift=0em]f1) {$\hat{Y}$};

    \draw[->, line width=\lw pt] (s1) -- (x1);
    \draw[->, line width=\lw pt] (y1) -- (x1);
    \draw[->, line width=\lw pt] (x1) -- (f1);
    \draw[->, line width=\lw pt] (f1) -- (yhat1);

    \node[align=center, anchor=north, scale=1.7] at ([xshift=2.5em, yshift=-35]x) {(a)};
    \node[align=center, anchor=north, scale=1.7] at ([xshift=2.5em, yshift=-35]x1) {(b)};
    
  \end{tikzpicture}    
    \caption{Dependence graphs~\cite{dehdashtian2024fairerclip} illustrating how biases from (a) inherent relations and (b) spurious correlations arise.\label{fig:causal-graph}}
    \vspace{-1.5em}
\end{figure}
From a \emph{machine learning perspective}, biases can be understood based on dependencies between data attributes, as illustrated in Fig.~\ref{fig:causal-graph}. The data $X$ (e.g., face images) depends both on the target attribute $Y$ (e.g., identity) and a sensitive attribute $S$ (e.g., skin tone) that induces bias. The goal of bias mitigation is to ensure that the prediction $\hat{Y}$ is statistically independent of $S$. The biases can be grouped into those arising from two scenarios: (1) $Y$ and $S$ are inherently dependent (\cref{fig:causal-graph}a). We refer to this type of correlation as \emph{intrinsic dependence}. (2) $Y$ and $S$ are independent (\cref{fig:causal-graph} b). In this case, we refer to any observed correlation as a \emph{spurious correlation}. In the latter case, we expect a bias-free model's performance w.r.t. $Y$ to be independent of $S$. However, the same is not so in the former case, where there will necessarily be a trade-off between performance and fairness.

Beyond the biases in the dataset, model design choices such as the objective function optimized during training, the sampling process used during training~\cite{li2019repair}, neural network architecture, etc., also have an influence on biases in the model predictions. These choices can either amplify or mitigate the biases in the training data. This is evidenced in the fact that models trained from the same biased training data can be made more or less biased by bias mitigation strategies~\cite{zhao2017men,wang2019balanced,wang2020towards,hendricks2018women}.

To understand the \emph{impacts} of biases, it helps to separate \emph{demographic biases} from \emph{non-demographic biases}. Demographic biases occur when models behave differently for different demographic groups. These groups can be defined in many ways and are usually specified by a protected attribute, such as \emph{gender, race,} or \emph{age}, among several others. Ideally, we expect the task performance of a bias-free model to be independent of such attributes. This reflects the goal of producing computer vision systems that are fair, inclusive, and equitable across segments of the world population. For example, the face recognition system of Fig.~\ref{fig:overview} should not be more accurate for lighter than darker faces. Non-demographic biases are not related to such demographic issues. For example, a person re-identification system can perform very effectively on certain datasets by simply matching clothing. However, this is only an illusion of good performance, as such systems will not be able to match people across images collected on different days. In this case, biases are spurious correlations that the computer vision systems learn to solve the task. These biases are not necessarily w.r.t. a known attribute, even though such attributes can be identified for many tasks, e.g. the \emph{clothing} attribute for re-identification, or the \emph{scene} and \emph{context} attributes for all recognition problems. Demographic and non-demographic biases are quite similar in the sense that they tend to originate from dataset or model biases and can be mitigated by similar algorithms. Hence, in what follows, we cover the two types of biases without much differentiation.

\subsection{Fairness Definitions}
\label{sec:fairnessdefinitions}
Multiple definitions of fairness~\cite{castelnovo2022clarification} have originated from social studies. Next, we describe the primary definitions of fairness in the computer vision literature.

\subsubsection{Individual Fairness}
Individual fairness seeks to ``treat similar individuals similarly" \cite{fleisher2021s}. One of the first attempts to formulate this objective was made by \cite{dwork2012fairness}, where Lipschitz conditions were employed. According to this condition, \emph{a small distance in feature space must translate to a small change in the model's decision}. The objective is defined as
\begin{equation}
    dist(\hat{y}_i, \hat{y}_j) < L \cdot dist(z_i, z_j)
\end{equation}
where $\hat{y}_i$ and $\hat{y}_j$ are decisions of the model, $z_i=f(x_i)$ and $z_j=f(z_j)$ are features of sample $i$ and $j$, respectively, and $L$ is the Lipschitz constant.

\subsubsection{Group Fairness}
Group fairness requires the model's decisions to be independent or conditionally independent of a sensitive (group) attribute. For example, university admissions' approval or rejection decisions must be independent of the applicant's \emph{gender}. In this example, the sensitive (group) attribute is \emph{gender}. There are three main definitions of group fairness: Demographic Parity (DP), Equal Opportunity (EO), and Equality of Odds (EOO). DP is defined as
\begin{align}
\scalemath{1}{
    P(\hat{Y} = y | S = s) = P(\hat{Y} = y | S = s') \nonumber} \\  
\scalemath{1}{
    \forall s, s' \in S, \forall y \in Y
    }
\end{align}
The university admissions example requires that the acceptance probability be equal for all genders. Although DP is a popular fairness definition, some studies ~\cite{hardt2016equality,dehdashtian2024utilityfairness} have argued that it is not practically relevant since it does not consider the true target label $Y$ for the decision. This problem is addressed by the other two definitions.

Equal opportunity (EO) is defined as
\begin{align}
\scalemath{0.95}{
    P(\hat{Y} = y | Y = y, S = s) = P(\hat{Y} = y | Y = y, S = s') \nonumber} \\ 
\scalemath{0.95}{    
    \forall s, s' \in S, \forall y \in Y
    }
\end{align}
In the university example, EO requires that the acceptance probability must be equal for all \emph{eligible} applicants from different sensitive groups. Finally, equality of odds (EOO) requires equal probability for mistakenly classifying accepted applicants from different sensitive groups as accepted applicants. It is formally defined as,
\begin{align}
\scalemath{0.95}{
    P(\hat{Y} = y_1 | Y = y_2, S = s) = P(\hat{Y} = y_1 | Y = y_2, S = s') \nonumber} \\ 
\scalemath{0.95}{
    \forall s, s' \in S, \forall y_1, y_2 \in Y
}
\end{align}
\subsubsection{Counterfactual Fairness}
Counterfactual fairness, defined by \citet{kusner2017counterfactual}, requires identical decision probabilities for a sample and its counterfactual counterpart. It requires intervention on sensitive attributes to not change the distribution of the model's decision \cite{zuo2023counterfactually}. It is formally defined as,
\begin{eqnarray}
\scalemath{0.8}{
    P\left( \hat{Y}_{S \leftarrow s} (U) = y | X = x, S = s \right) = P\left( \hat{Y}_{S \leftarrow s'} (U) = y | X = x, S = s \right)  \nonumber} \\ 
\scalemath{0.8}{
    \forall s, s' \in S
    }
\end{eqnarray}
\noindent where $U$ is an unobserved variable in the causal graph (\cref{fig:causal-graph}). In the university's admission example, if the model accepts a \emph{male} applicant, it should make the same decision if the applicant were \emph{female}, assuming all other attributes are adjusted accordingly. Note that counterfactual samples are not created merely by changing the sensitive attribute; instead, they are generated by considering the changes in other attributes that result from the alteration of the sensitive attribute due to causal relationships between them.

\subsubsection{Bias Amplification}
\label{sec:biasamplification}
Another phenomenon studied in bias quantification is the exacerbation of biases beyond those present in the dataset during model training. This is usually called {\em bias amplification}~\cite{zhao2017men,wang2021directional}. It is understood as the difference in the biases exhibited by a trained machine learning model relative to the biases present in the data used to train such a model. This term, first used in \citet{zhao2017men} for the task of situation recognition, has been used to provide a notion of bias that does not depend on any pre-existing notion of fairness w.r.t. parity. Reducing bias amplification in a model is equivalent to reducing the biases only to the extent to which they are already present in the training set. For instance, a model that predicts a label at disparate rates for people of different genders will only be considered to suffer from bias amplification if the rates are different from those present in the training data. \citet{wang2021directional} define the notion of {\em directional bias amplification}, which further refines the bias amplification measure by accounting for varying base rates of the protected attributes.

\section{Bias Discovery and Analysis \label{sec:analysis}}

This section discusses a series of works that discover or analyze biases in computer vision datasets and models. The goal is to identify inherent biases that threaten fairness and generalization. Uncovering such biases raises awareness of potential limitations and biases in computer vision systems and helps guide future work to develop more equitable and robust computer vision systems.

\subsection{Biases in Datasets}
Dataset biases can propagate to computer vision models or get amplified by the models, thereby influencing their fairness and performance. We review studies that analyze biases in commonly used datasets. %

\citet{meister2023gender} delve into gender biases in large-scale visual datasets, exploring how gender information can be removed from datasets and how visual cues, or ``gender artifacts", influence model predictions. \citet{guilbeault2024online} compare gender biases in images and text across massive online corpora, revealing how visual content may amplify gender biases more than textual content. \citet{shankar2017no} investigate the geographical diversity of large datasets such as ImageNet~\cite{deng2009imagenet} and Open Images~\cite{krasin2017openimages}, revealing noticeable Amerocentric and Eurocentric biases that affect model performance across different global regions. In the context of facial image datasets \citet{chen2021understanding} found that significant gender biases were introduced in the annotations, especially related to facial expressions. More recently, in the context of foundation models, \citet{birhane2023into,birhane2023hate} examined the presence of hate content in the text annotations of LAION~\cite{schuhmann2022laion}, a large-scale dataset commonly used for training vision-language models. They found significant levels of hate content which increased by 12.26\% when scaling from LAION-400M to LAION-2B.

\subsection{Biases in Models}
Biases in computer vision models can impact their performance and fairness, especially when these models are deployed in the real world. Numerous studies sought to identify the presence and impact of biases in different types of learning methods and pretrained models across a diverse set of visual recognition tasks.

In their pioneering work, \citet{buolamwini2018gender} evaluated and uncovered gender and skin-tone biases in many commercial face recognition systems and highlighted serious implications for high-stakes contexts such as healthcare and law enforcement. \citet{domnich2021responsible} investigated bias w.r.t. gender in various facial expression recognition models, assessing and identifying which architectures and emotions are more influenced by gender.

\citet{sirotkin2022study} investigated the origins and impact of social biases in self-supervised learning (SSL) methods, revealing the correlations between the different SSL algorithms and the number of inherent biases. \citet{iofinova2023bias} examine how model compression algorithms like pruning induce or exacerbate biases, particularly affecting marginalized groups by increasing systematic and category biases under high sparsity levels. Analysis on vision transformers~\cite{brinkmann2023multidimensional} measured factors contributing to social biases by investigating training data, objectives, and architectures. \citet{wilson2019predictive} explore the performance disparities of object detection models in autonomous driving, explicitly revealing poorer detection rates for pedestrians with Fitzpatrick skin types 4 to 6, and investigate contributing factors such as training set composition, measurement issues, and the impact of loss function prioritization.

A few studies proposed approaches to audit computer vision models for potential biases. \citet{ranjit2023variation} propose a framework to audit and analyze pretrained visual recognition models for biases w.r.t. sensitive visual attributes, evaluating how these biases change after fine-tuning. Studies on biases of pretrained models on downstream tasks show that such models can inherit biases related to spurious correlations and underrepresentation, but these biases can be mitigated by carefully curating the finetuning dataset~\cite{wang2023overwriting}. Similarly, \citet{birhane2024dark} found that transformer-based CLIP models inherited racial biases prevalent in the LAION dataset on which they were trained. 

Concurrently, \citet{sadeghi2022on} and \citet{dehdashtian2024utilityfairness} defined and estimated the near-optimal trade-offs between model performance (accuracy of predicting target attributes) and different group fairness definitions. These trade-offs were utilized to evaluate (a) more than 100 pre-trained CLIP models from OpenCLIP~\cite{ilharco_gabriel_2021_5143773}, (b) more than 900 pre-trained image models from Pytorch Image Models~\cite{rw2019timm}, and (c) existing fair representation methods on CelebA and FairFace datasets. The results (shown in Fig.\ref{fig:trade-offs}) revealed that, out of the box, pre-trained models were far from the best achievable limits of performance and fairness, thus identifying the significant limitations of existing computer vision models and the dire need for further research to make computer vision systems more socially responsible and equitable. Furthermore, such an evaluation can also help the community identify trends and pre-trained models that best suit their specific task and dataset.

\subsection{Biases Beyond Demographic Attributes}
Various forms of biases beyond demographic attributes have also been studied in computer vision datasets and models. \citet{torralba2011unbiased} first introduce the notion of \emph{dataset bias}, and identify the distribution gaps between different vision datasets w.r.t. viewpoints, styles, and scenes. \citet{geirhos2018imagenet} discover and analyze the \emph{texture bias} in CNN object classifiers, finding the models more sensitive to local textures while overlooking object shapes. \citet{li2018resound} study the \emph{representation bias} in action recognition datasets, in which action labels are implied through scenes and background objects. \citet{zhang2016yin} investigate \emph{unimodal bias} in VQA datasets and show that many visual questions can be answered correctly by using language prior alone.

These studies pave the way for the collection of new, bias-controlled datasets \cite{li2018resound,geirhos2018imagenet,agrawal2018don}, either to evaluate the models under an unbiased setting or as training data to remedy model bias. They also offer insights into how vision models inherit biases from the data, and as discussed next leading to various mitigation methods for training models less susceptible to biases~\cite{khosla2012undoing,choi2019can,cadene2019rubi,sadeghi2019global,bahng2020learning,sadeghi2022on,dehdashtian2024fairerclip,dehdashtian2024utilityfairness}.

\section{Bias Mitigation Methods \label{sec:method}}
This section summarizes common approaches to mitigate bias across various tasks. Each subsection is dedicated to a specific category of algorithms, detailing their applications. By organizing the algorithms this way, we aim to provide a clear understanding of bias mitigation in computer vision.

    \begin{table*}%
    \renewcommand{\arraystretch}{1.2} %
    
    \caption{Bias analysis and mitigation for vision and language models\label{tab:methods-vl-table}} \label{tab:methods-vl}
    \begin{tabular}{p{3cm}p{3cm}p{10.4cm}}
    
    \toprule
    Task & Attribute & References \\
    \midrule

    \multirow{3}{*}{Image Captioning} & 
    Gender & \citet{burns2018women}, \citet{bhargava2019exposing}, \citet{hirota2023model} \\
    & Race & \citet{zhao2021understanding} \\
    & Social & \citet{hirota2022quantifying} \\
    \midrule

    \multirow{12}{*}{Text-to-Image Synthesis} & 
    Gender &
    \citet{esposito2023mitigating},
    \citet{friedrich2023fair},
    \citet{he2024debiasing},
    \citet{luccioni2024stable},
    \citet{cho2023dall} \\
    
     & Race &
     \citet{esposito2023mitigating}, 
     \citet{friedrich2023fair},
     \citet{bansal2022well},
     \citet{he2024debiasing},
     \citet{luccioni2024stable}\\
     
    & Adjective &\citet{luccioni2024stable} \\
    & Profession &\citet{wang2023t2iat}, \citet{luccioni2024stable}, \citet{cho2023dall} \\
    & General &\citet{chinchure2023tibet}, \citet{zhang2023iti} \\
    & Stereotype &\citet{bianchi2023easily} \\
    & Pose &\citet{ruiz2023dreambooth} \\
    & Culture &\citet{liu2024scoft} \\
    & Geography &\citet{basu2023inspecting} \\
    & Skin Tone &\citet{cho2023dall} \\
    \midrule
  
    \multirow{6}{*}{Question Answering} & 
    Language &
    \citet{manjunatha2019explicit},
    \citet{kv2020reducing},
    \citet{niu2021counterfactual},
    \citet{kervadec2021roses},
    \citet{dancette2021beyond},
    \citet{wen2021debiased},
    \citet{cho2023generative},
    \citet{basu2023rmlvqa} \\
    
    & Gender &
    \citet{park2020fair},
    \citet{hirota2022gender} \\
    & Visual Context &
    \citet{selvaraju2020squinting} \\
    & Correlations &\citet{agarwal2020towards}, \citet{gupta2022swapmix} \\
    & Race &
    \citet{hirota2022gender} \\
    \midrule
    
    \multirow{3}{*}{CLIP De-biasing} 
    & Gender & 
    \citet{dehdashtian2024fairerclip}, \citet{seth2023dear}, \citet{chuang2023debiasing}, \citet{berg2022prompt}, \citet{alabdulmohsin2024clip}\\
    & Race & 
    \citet{dehdashtian2024fairerclip}, \citet{berg2022prompt}, \citet{chuang2023debiasing}, \citet{seth2023dear}\\
    & Background & 
    \citet{dehdashtian2024fairerclip}, \citet{chuang2023debiasing}, \citet{phan2024controllable} \\
    
    \midrule
    \multirow{2}{*}{Other} 
    & \multirow{2}{*}{-} & 
    \citet{kong2024mitigating},
    \citet{yenamandra2023facts},
    \citet{qiu2023cafeboost},
    \citet{shankar2017no},
    \citet{chu2021learning},
    \citet{garcia2023uncurated},
    \citet{biswas2023probabilistic},
    \citet{tang2020unbiased},
    \citet{cui2023holistic} \\
    \bottomrule
    \end{tabular}
    \end{table*}

\subsection{Fairness through Unawareness}
A naive approach to fairness is to withdraw sensitive protected attributes from data or to avoid those protected attributes as input to the machine learning model. This is often referred to as fairness through {\em blindness} or fairness through unawareness. It has been well documented in the machine learning literature that this approach is frequently ineffective. As discussed above, correlations between sensitive protected attributes and other attributes can still lead to biases in model predictions, e.g., zip codes being informative of race in the case of credit scoring systems. 

In computer vision, fairness through unawareness could be attempted by blurring people's faces or removing people entirely from images. Such an approach parallels the drawbacks observed in the machine learning literature. For instance, background pixels that are not blurred or obscured may correlate highly with sensitive attributes, e.g.~men more commonly wear ties and women more commonly wear dresses, so clothing will correlate highly with gender. Another drawback is that for many vision tasks, such as human activity recognition, the visual features of people are essential to accomplish the task accurately. Hence, blurring or obscuring people interferes with this goal. \citet{wang2019balanced} used fairness through unawareness as a baseline and demonstrated that adversarial bias mitigation by explicitly modeling the sensitive attribute leads to better outcomes.

\begin{figure}
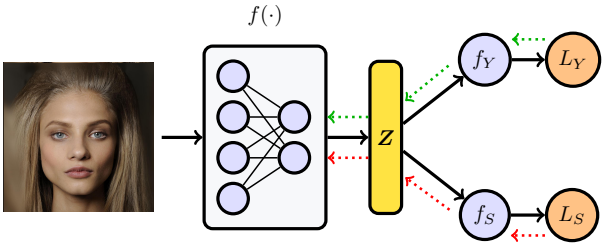

    \centering
    \hspace{-3em}\includestandalone[width=\linewidth]{figs/adversarial}
    \caption{\textbf{Fair Representation Learning.} An encoder $f$ maps images to a representation $Z$. A target branch maximizes the statistical dependence between $Z$ and $Y$, while a fairness branch minimizes the statistical dependence between $Z$ and the protected attribute $S$. Methods in this class differ in the choice of loss functions $L_Y$, $L_S$, models for $f_Y$ and $f_S$, and learning (iterative vs closed-form, local vs global optima).\label{fig:adv-model}}
    \vspace{-1.5em}
\end{figure}
\subsection{Fair Representation Learning}
Over the last decade, several approaches have been developed for learning fair image representations (see Figure~\ref{fig:adv-model} for an illustration). These approaches follow the template of adopting a fairness constraint (e.g., $Z\indep S$ for demographic parity or $Z\indep S | Y=y$ for Equality of Odds) as a regularizer in addition to the objective for the target task. The approaches differ in two respects: (a) the choice of measure as a proxy for quantifying the statistical dependence corresponding to $Z\indep S$ and $Z\indep S | Y=y$, and (b) the associated optimization technique.

From a \emph{proxy dependence measure} perspective, existing approaches either measure (i) the degree of linear dependence between $Y$ and $Z$, (ii) the degree of mean dependence, i.e., $\mathbb{E}(Z)\indep S$ or matching only the first moment of the distribution, or (iii) the degree of full statistical dependence, i.e., matching all moments of the distribution. Adversarial representation learning (ARL)~\cite{edwards2015censoring,xie2017controllable,madras2018learning,wang2019balanced,roy2019mitigating} adopts neural network-based classifiers or regressors as a proxy measure of statistical dependence between $Z$ and $S$, which is equivalent to mean dependence~\cite{adeli2021representation} only. State-of-the-art approaches~\cite{quadrianto2019discovering, sadeghi2022on, dehdashtian2024utilityfairness, dehdashtian2024fairerclip}, however, adopt non-parametric independence measures like the Hilbert-Schmidt Independence Criterion (HSIC)~\cite{gretton2005measuring} and its variants~\cite{sadeghi2022on}, which measures full statistical dependence and can enforce independence-based fairness constraints more effectively.

From an \emph{optimization} perspective, solutions have either adopted iterative approaches like two-player zero-sum min-max optimization for ARL~\cite{xie2017controllable} that converge to local optima or closed-form solvers~\cite{sadeghi2019global, sadeghi2022on} that lead to global optima of the underlying optimization. Due to the inherent instability of zero-sum min-max optimization, several variants of ARL have been proposed. \citet{roy2019mitigating} proposed a non-zero-sum variant of ARL to improve the convergence properties of the ARL optimization. \citet{sadeghi2019global} studied ARL from an optimization perspective and obtained a closed-form solution that affords global optima of the ARL optimization through spectral learning and provided theoretical guarantees for achieving utility and fairness. \citet{sadeghi2021adversarial} used a kernel ridge regressor to model the adversary and backpropagated through its closed-form solution, resulting in stable optimization and improved performance utility-fairness trade-off. \citet{sadeghi2022on} proposed a non-parametric dependence measure to capture all non-linear statistical dependencies and obtained a global optimum of the underlying optimization problem through a closed-form solution, thus obtaining provably near-optimal utility-fairness trade-offs.

A majority of the debiasing methods in computer vision are based on ARL. For instance, cross-sample adversarial debiasing (CSAD) disentangles target and bias features to prevent biased decision-making~\cite{zhu2021learning}. The Causal Attention Module (CaaM) employs an adversarial minimax fashion to disentangle and optimize complementary attention mechanisms~\cite{wang2021causal}.  The Lottery Ticket Hypothesis is adopted to find fair and accurate subnetworks without weight training, leveraging fairness regularization and adversarial training for bias mitigation~\cite{tang2023fair}. Furthermore, fairness-aware adversarial perturbation (FAAP) modifies input data to conceal fairness-related features from deployed models without adjusting the model parameters or structures~\cite{wang2022fairness}. FAIRREPROGRAM introduces fairness triggers appended to inputs, optimizing them under a min-max formulation with an adversarial loss to obscure demographic biases~\cite{zhang2022fairness}.

In face recognition, adversarial learning reduces the encoding of sensitive attributes in face representations. For instance, adversarial learning frameworks can disentangle demographic information from identity representations, reducing bias in face recognition and demographic attribute estimation~\cite{gong2020jointly}. The Protected Attribute Suppression System (PASS) employs a discriminator to prevent networks from embedding protected attribute information, thereby mitigating gender and skin tone biases without end-to-end training~\cite{dhar2021pass}. Techniques using the Hilbert-Schmidt independence criterion transform input data into fair representations that maintain semantic meaning while ensuring statistical independence from protected characteristics~\cite{liu2019fair}. For generative models, adversarial methods harmonize adversarial training with reconstructive generation to improve data coverage and include minority groups more effectively~\cite{yu2020inclusive}. Lastly, unknown biased attributes in classifiers can be identified by optimizing a hyperplane in a generative model's latent space using total variation loss and orthogonalization penalty~\cite{li2021discover}.

\subsection{Accuracy-Unfairness Trade-Offs \label{sec:method:trade-off}}
In scenarios where the target attribute $Y$ and the sensitive attribute $S$ exhibit considerable statistical dependency, the objectives of learning a fair representation—specifically, removing information related to $S$ while retaining information pertinent to $Y$—are in conflict. This conflict impacts the performance of these objectives. Consequently, a trade-off exists between the retention of $Y$-related information and the removal of $S$-related information. This trade-off can be observed through the accuracy, MSE loss, F1 score, etc., of predicting $Y$ and the fairness of the decisions made by the model. We generally use the word \emph{utility} to refer to the model's performance in retaining $Y$-related information. 
\begin{figure}[t]
    \centering
    \begin{subfigure}{0.24\textwidth}
        \centering
        \includegraphics[width=0.73\textwidth]{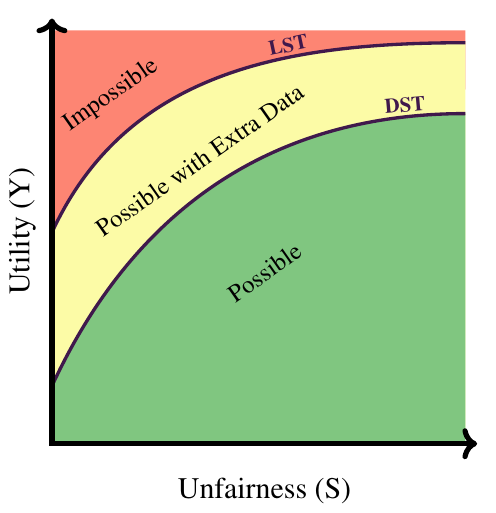}
    \end{subfigure}
    \begin{subfigure}{0.24\textwidth}
        \centering
        \includegraphics[width=0.9\textwidth]{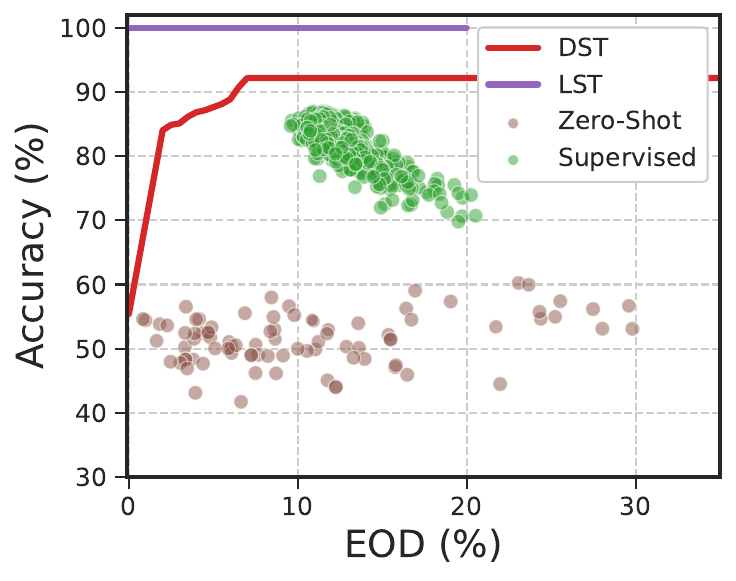}
    \end{subfigure}
    \caption{\textbf{The utility-fairness trade-offs.}\cite{dehdashtian2024utilityfairness} (Left) Models can be evaluated by their utility (e.g., accuracy, MSE loss, F1 score, etc.) w.r.t. a target label $Y$ and their unfairness w.r.t. a sensitive attribute $S$. \citet{dehdashtian2024utilityfairness} introduce two trade-offs, \emph{Data Space Trade-Off} (DST) and \emph{Label Space Trade-Off} (LST). (Right) \citet{dehdashtian2024utilityfairness} empirically estimate DST and LST on CelebA and evaluate the utility (high cheekbones) and fairness (gender \& age) of over 100 zero-shot and 900 supervised image models. \label{fig:trade-offs}}
    \vspace{-1.5em}
\end{figure}

The existence of a utility-fairness trade-off has been well established both theoretically \cite{sadeghi2022on,menon2018cost,zhao2022inherent,wang2024aleatoric} and empirically \cite{sadeghi2022on, dehdashtian2024utilityfairness}. \citet{sadeghi2022on} characterize the near-optimal trade-off for multidimensional continuous/discrete attributes using a closed-form solution on the extracted features from a frozen feature extractor. Additionally, \citet{dehdashtian2024utilityfairness} define two trade-offs: the Data Space Trade-Off (DST) and the Label Space Trade-Off (LST). These trade-offs capture the intrinsic relationship between the $Y$ and $S$ labels independently of the samples. They achieve this by employing a trainable feature extractor alongside a closed-form solution for the fair encoder. Although these studies focus on estimating trade-offs, the methods proposed can also serve as state-of-the-art bias mitigation techniques, as they identify and operate within the best achievable regions.

Estimating these trade-offs can be beneficial in several ways \cite{dehdashtian2024utilityfairness}: (1) they provide users with more information to make informed decisions when choosing a pre-trained model, and (2) they illustrate the extent to which fine-tuning can improve the utility or fairness of the model. \Cref{fig:trade-offs} illustrates the plausible trade-offs, their empirical estimation on CelebA~\cite{liu2015deep}, and their utility in empirically evaluating representations from pre-trained models.

\subsection{Counterfactual Data Rebalancing}
Counterfactual data rebalancing addresses bias by generating or reweighting data to create balanced representations of different groups. Several approaches have been proposed to operationalize this conceptual idea. In image classification, GAN-based data augmentation has been combined with adaptive sampling for disadvantaged group accuracy enhancement~\cite{zietlow2022leveling}. FlowAug employs flow-based generative models to create semantically augmented images, reducing subgroup performance discrepancies by addressing spurious correlations~\cite{chiu2023better}. The Confidence-based Group Label assignment (CGL) method assigns pseudo group labels to unlabeled samples based on prediction confidence~\cite{jung2022learning}.

In face recognition, methods like INV-REG self-annotate demographic attributes and impose invariant regularization during training to learn causal features robust across diverse demographic groups~\cite{ma2023invariant}. StyleGANs transfer multiple demographic attributes simultaneously, enhancing dataset diversity and mitigating bias in face recognition systems~\cite{georgopoulos2021mitigating}. It is also possible to generate synthetic images to supplement imbalanced datasets, creating a semi-synthetic balanced dataset to improve fairness in facial attribute and gender classification tasks~\cite{li2022cat}.

In semantic segmentation, randomly dropping class-specific feature maps disentangles class representations and mitigates dataset biases~\cite{chu2021learning}.
In image captioning, a Debiasing Caption Generator (DCG) has been proposed to correct gender-biased captions, forming a model-agnostic debiasing framework~\cite{hirota2023model}.
In action recognition, StillMix mitigates background and foreground static bias by mixing bias-inducing frames with training videos~\cite{li2023mitigating}.
Person Re-identification methods introduce causal inference to eliminate clothing bias from identity representation learning~\cite{yang2023good}.

In scene graph generation, counterfactual causality isolates and removes the effects of context bias~\cite{tang2020unbiased}.
In federated learning, Bias-Eliminating Augmenters (BEA) at each client generate bias-conflicting samples, thereby mitigating local data biases during distributed training~\cite{xu2023bias}.
For bias discovery, model reliance on spurious correlations is amplified to segregate bias-conflicting samples, which are then identified and mitigated through a slicing strategy~\cite{yenamandra2023facts}.

In representation learning, biases can be mitigated through resampling, penalizing examples that are easily classified by a specific feature representation, and reweighting the dataset through a minimax optimization problem~\cite{li2019repair}. Methods like BiaSwap create balanced datasets through an unsupervised image translation-based augmentation framework that identifies and swaps bias-relevant regions in images~\cite{kim2021biaswap}. Some methods use GANs to generate images with altered combinations of target and protected attributes to decorrelate them~\cite{ramaswamy2021fair}. Creating bias-reducing positive and negative sample pairs from a self-supervised object localization method can also mitigate scene bias~\cite{mo2021object}. A fair contrastive learning method uses gradient-based reweighting to learn fair representations without demographic information by incorporating a small labeled set into the self-supervised training process~\cite{chai2022self}. Other methods identify bias pseudo-attributes via clustering and reweight these clusters based on their size and task-specific accuracy to improve worst-group generalization~\cite{seo2022unsupervised}. Some approaches identify intermediate attribute samples near decision boundaries and use them for conditional attribute interpolation to learn debiased representations~\cite{zhang2023learning}. Class-conditioned sampling mitigates bias by creating multiple balanced dataset variants, each with a subsampled distribution that mimics the bias distribution of the target class~\cite{qraitem2023bias}. The Debiased Contrastive Weight Pruning (DCWP) method identifies bias-conflicting samples and uses them to train a pruned neural network~\cite{park2023training}.

\subsection{Score Calibration and Loss Regularization} %
Score calibration adjusts the decision thresholds or prediction scores of models to ensure fair outcomes across different demographic groups. In image classification, score calibration can mitigate contextual bias by minimizing overlap in class activation maps and learning uncorrelated feature representations to ensure accurate recognition of both in and out of typical category contexts~\cite{singh2020don}. Some methods propose fairness surrogates to optimize constraints in network training~\cite{bendekgey2021scalable}, while others address spurious correlations and intrinsic dependencies with non-parametric measures of statistical dependence~\cite{dehdashtian2024fairerclip}. U-FaTE~\cite{dehdashtian2024utilityfairness} quantifies utility-fairness trade-offs by optimizing a weighted combination of statistical dependence measures to evaluate and improve the fairness of pre-trained models.

In face recognition, a fair loss with an adaptive margin strategy optimized via reinforcement learning has been proposed to address class imbalance~\cite{liu2019fair}. The Group Adaptive Classifier (GAC) uses adaptive convolution kernels and channel-wise attention maps to learn general and demographic-specific patterns and reduce demographic bias in face recognition~\cite{gong2021mitigating}. 

An Equalizer Model with two complementary losses has been proposed for image captioning to leverage gender-specific visual evidence and generate gender-neutral words when uncertain~\cite{hendricks2018women}. To achieve equal representation in the image retrieval task, a test-time post-processing algorithm creates fair retrieval subsets by using predicted gender or race attributes from the classifier or zero-shot inference~\cite{kong2024mitigating}. For Bayesian networks, posterior inference can combine within-triplet priors with uncertain evidence to mitigate long-tailed bias~\cite{biswas2023probabilistic}. In continual learning, task-induced bias can be mitigated using causal interventions with attention mechanisms that transform biased features into unbiased features~\cite{qiu2023cafeboost}.

In representation learning, a regularization strategy entangles features from the same target class and disentangles biased features~\cite{tartaglione2021end}. Fairness-aware feature distillation improves fairness using maximum mean discrepancy to align the distributions of group-conditioned features from a student model with the group-averaged features of a teacher model~\cite{jung2021fair}. A fair contrastive loss and a group-wise normalization are proposed in ~\cite{park2022fair} to prevent the inclusion of sensitive attribute information and balance loss based on group cardinality, respectively. Leveraging hierarchical features and orthogonal regularization has also been shown to mitigate unknown biases~\cite{jeon2022conservative}.

\section{Datasets \label{sec:datasets}}

In this section, we summarize various datasets used in fairness-related tasks in computer vision along with their corresponding sensitive attributes. The tasks and their datasets are listed in \Cref{tab:attr-table}. The tasks range from action recognition and text-to-image to face recognition and classification. We also discuss datasets used in multiple tasks, those specialized for a single task, and the attributes investigated most and least.

\subsection{Diversity of Datasets and Attributes}
Some datasets in the table are used across multiple fairness-related computer vision tasks. For instance, MSCOCO~\cite{lin2014microsoft} and its variants \cite{caesar2018cvpr,hendricks2018women} are used in bias analysis and evaluation of fairness in the context of classification, image captioning, image retrieval, scene graph generation, and VQA. Similarly, CelebA~\cite{liu2015deep} is extensively used in evaluating and mitigating bias for classification, face recognition, and text-to-image tasks, emphasizing the need to address biases in gender, ethnicity, and age. UTKFace~\cite{zhang2017age} is employed in fairness in classification, face recognition, and representation learning for investigating and mitigating biases related to \emph{age}, \emph{gender}, \emph{ethnicity}, and \emph{skin tone}. OpenImages~\cite{krasin2017openimages} is also used in both bias analysis and representation learning with a focus on \emph{gender} and \emph{geography} biases. The repeated use of these datasets underscores their importance and highlights their value in providing diverse annotations for evaluating and mitigating bias across computer vision tasks.

It is evident from analyzing \Cref{tab:task-datasets} that certain sensitive attributes are more frequently investigated across various computer vision tasks, while others receive less attention. \emph{Gender} stands out as the most frequently studied attribute across tasks such as bias analysis, classification, face recognition and analysis, image captioning, image retrieval, representation learning, text-to-image, and VQA. Datasets like MSCOCO~\cite{lin2014microsoft,caesar2018cvpr,hendricks2018women}, CelebA~\cite{liu2015deep}, OpenImages~\cite{krasin2017openimages}, IMDB Face~\cite{wang2018devil}, FairFace~\cite{karkkainen2019fairface}, and UTKFace~\cite{zhang2017age} are often used to explore and mitigate \emph{gender} biases. This reflects the bias of the computer vision community towards evaluating and mitigating \emph{gender} biases.

\emph{Race} and \emph{ethnicity} are also critical attributes, especially in face recognition and representation learning tasks. Datasets such as PPB~\cite{buolamwini2018gender}, IJB-A~\cite{klare2015pushing}, Fairface~\cite{karkkainen2019fairface}, UTKFace~\cite{zhang2017age}, MSCOCO~\cite{lin2014microsoft}, and Adience~\cite{adience} are frequently used to investigate these biases. Similarly, \emph{age} is a commonly investigated attribute in classification, face recognition, and representation learning tasks with datasets like IMDB Face~\cite{rothe2018deep}, UTKFace~\cite{zhang2017age}, MORPH~\cite{ricanek2006morph}, CACD~\cite{chen2014cross}, and MS-Celeb-1M~\cite{gong2020jointly, deng2019arcface} being used to examine \emph{age}-related biases.

\emph{Context} is another frequently considered attribute, especially in classification, image captioning, representation learning, and VQA. This attribute is investigated using datasets like MSCOCO~\cite{lin2014microsoft}, UnRel~\cite{Peyre17}, Deep Fashion~\cite{liu2016deepfashion}, NICO++~\cite{zhang2023nico++}, and PATA~\cite{seth2023dear}.

In contrast to the above-mentioned attributes, certain attributes have received less attention. For example, the \emph{illumination} attribute is only addressed by the Extended Yale B~\cite{Extended-Yale-B} dataset within the classification task. Similarly, \emph{Hair Color} is primarily referenced in the CelebA dataset. While \emph{Skin Tone} is addressed in some tasks, it appears less frequently compared to attributes like \emph{gender} or \emph{race}. Additionally, \emph{Texture} is an attribute less commonly investigated, appearing mainly in the ImageNet-A~\cite{hendrycks2021natural} dataset for representation learning. Lastly, \emph{corruption} is mainly mentioned in the context of federated learning and representation learning in datasets like Corrupted \mbox{CIFAR-10}~\cite{hendrycks2018benchmarking}. 

\subsection{Task Specific Diversity of Datasets and Attributes}
Underscoring the need for a more detailed examination of biases in specific domains, a diverse array of datasets and attributes have been specialized for each domain.

The \textbf{action recognition} task exhibits a moderate diversity in its datasets, incorporating a range of scene contexts from UCF-101~\cite{soomro2012ucf101} and HMDB-51~\cite{HMDB} to specialized datasets like Diving48~\cite{li2018resound} and THUMOS-14~\cite{THUMOS14}. These datasets provide various environments and activities, ensuring varied training and evaluation conditions. However, the sensitive attribute used in this task is \emph{scene}, as biases can arise from the background, objects, or context in which action occurs. This emphasis on scene-based attributes highlights the need to mitigate biases that stem from environmental contexts affecting action recognition performance. Without such mitigation, computer vision models can leverage context as a shortcut to solve the action recognition problem without understanding any action~\cite{li2018resound}.

Papers in the \textbf{bias analysis} task leverage a diverse array of datasets, including MSCOCO~\cite{lin2014microsoft}, OpenImages~\cite{krasin2017openimages}, and CelebA~\cite{liu2015deep}, which cover multiple demographic and contextual attributes. The most commonly studied attributes in this task are \emph{gender} and \emph{age}, reflecting a significant concern within the computer vision community regarding the impact of these biases on model decisions.

Frequently studied attributes in \textbf{face recognition and analysis} studies include \emph{gender}, \emph{ethnicity/race}, \emph{skin tone}, and \emph{age}. These studies have been performed on a diverse set of datasets like PPB~\cite{buolamwini2018gender}, IJB-A~\cite{klare2015pushing}, and Adience~\cite{adience}.

\textbf{Federated learning} tasks use specialized datasets, such as Colored MNIST~\cite{arjovsky2019invariant} and Corrupted CIFAR-10~\cite{hendrycks2018benchmarking}, focusing on attributes like color and corruption. While the diversity of the attributes is limited compared to other tasks, they are tailored to study specific biases and robustness issues prevalent in federated learning environments.

\textbf{Image captioning} tasks utilize datasets like MSCOCO-Bias~\cite{hendricks2018women} and MSCOCO-Balanced~\cite{hendricks2018women}, which are designed to highlight and mitigate \emph{gender} and \emph{racial} biases in image descriptions, reflecting a desire for fair and representative captions. The diversity of these datasets lies in their annotations and the variety of contexts they provide.

The \textbf{image retrieval} task makes use of datasets such as Occupation 1~\cite{kay2015unequal}, Occupation 2~\cite{celis2020implicit}, MSCOCO~\cite{lin2014microsoft}, and Flickr30k~\cite{plummer2015flickr30k}, with a primary focus on \emph{gender}. 

Fairness studies in \textbf{object detection} utilize the BDD100K~\cite{BDD100K} dataset, focusing on attributes like \emph{skin tone} and \emph{income}. The diversity in this task is centered around addressing biases that are particularly relevant in autonomous driving and other detection-based applications.

\textbf{Person re-identification} tasks use datasets such as PRCC-ReID~\cite{yang2019person} and LTCC-ReID~\cite{qian2020long}, primarily focusing on the attribute of \emph{clothing}, highlighting the importance of mitigating biases w.r.t. clothing for this task.

\textbf{Representation learning} tasks demonstrate high diversity with datasets like ImageNet~\cite{russakovsky2015imagenet,shankar2017no}, Open Images~\cite{krasin2017openimages,shankar2017no}, MSCOCO~\cite{lin2014microsoft}, and CIFAR-10S~\cite{softLabelElicitingLearning2022}, addressing a broad range of attributes including \emph{geography}, \emph{gender}, \emph{color}, \emph{corruption}, \emph{scene}, and \emph{context}. The focus on multiple attributes indicates an effort to create robust models that generalize well across various demographic and environmental factors.

\textbf{Scene graph generation} employs datasets such as VG~\cite{krishna2017visual} and MSCOCO~\cite{lin2014microsoft}, with an emphasis on the attribute of \emph{composition}. This task's diversity in datasets is geared towards understanding and mitigating biases in scene understanding and object relationships.

For debiasing \textbf{text-to-image} models datasets like CelebA~\cite{liu2015deep}, FAIR~\cite{feng2022towards}, and FairFace~\cite{karkkainen2019fairface} are used. The primary focus in this task has been debiasing with respect to attributes such as \emph{gender} and \emph{skin tone}. By leveraging these diverse datasets, researchers aim to address biases that can arise from textual prompts influencing image generation. 

In \textbf{visual question answering} (VQA), a variety of datasets are utilized, including VQA~\cite{VQA}, MSCOCO~\cite{lin2014microsoft}, VQA-CP v2~\cite{agrawal2018don}, and Visual7W~\cite{zhu2016visual7w}. These datasets address attributes like \emph{language}, \emph{context}, \emph{gender}, and \emph{race}, ensuring comprehensive evaluation and mitigation of biases in multimodal understanding.

\section{Current Trends and Future Work \label{sec:finish}}

\noindent\textbf{Fairness in Generative Models:} The availability of large multimodal datasets \cite{schuhmann2022laion}, coupled with significant advancements in generative modeling \cite{esser2024scaling,sohl2015deep,ho2020denoising,ramesh2022hierarchical,ramesh2021zero,rombach2022high,zhang2023adding,podell2023sdxl}, has substantially increased the capabilities and potential applications of generative models~\cite{zhang2023iti}. Concurrently, some studies~\cite{bianchi2023easily,ramesh2022hierarchical,rombach2022high} began demonstrating that the content generated by these models exhibits biases. However, these have predominantly focused on the diversity of the generated images across different demographic groups. As such, a formal and precise mathematical definition of fairness still does not exist for generative models.

In the context of Text-to-Image (TTI) generation, approaches for increasing the diversity of generated content~\cite{jalal2021fairness,friedrich2023fair,luccioni2024stable} can be categorized into two main groups, namely prompt engineering and guidance. \underline{\emph{Prompt Engineering:}} Methods~\cite{wu2024contrastive} in this category focus on designing better text prompts to force the model to generate more diverse images. \underline{\emph{Guidance:}} These methods~\cite{dhariwal2021diffusion,liu2023more,nichol2021glide} seek to improve the diversity of generated images by modifying the distribution of generated images. However, all of these methods still struggle to achieve fine-grained control over the generation process, often resulting in unintended changes to other attributes when attempting to modify the protected ones. This challenge is primarily because different attributes are entangled with each other~\cite{wu2024contrastive}.

\vspace{2pt}
\noindent\textbf{Fairness in Foundation Models:} Recent advances in transformer architectures and large-scale pretraining have led to the development of families of multimodal foundation models~\cite{radford2021learning,alayrac2022flamingo,kirillov2023segment} that demonstrate remarkable generalization capacity to novel tasks and domains. Just like application-specific models, foundation models also exhibit demographic and other biases across different downstream tasks. This has been observed for tasks including classification~\cite{agarwal2021evaluating}, retrieval~\cite{berg2022prompt}, captioning~\cite{hirota2023model} and visual question answering~\cite{ruggeri2023multi}. Bias analysis and mitigation methods in this area mainly concern two types of foundation models.

\underline{\emph{Image-Text Models:}} Some efforts have been made to address bias in the image and text \emph{representations} learned by contrastive models such as CLIP~\cite{radford2021learning} and SigLIP~\cite{zhai2023sigmoid} in the context of zero-shot image classification and image-text retrieval tasks. These approaches typically employ a set of sensitive text or image queries to debias CLIP embeddings through prompt tuning~\cite{berg2022prompt}, auxiliary modules~\cite{zhang2022contrastive,seth2023dear}, and linear~\cite{chuang2023debiasing} or nonlinear~\cite{dehdashtian2024fairerclip} feature mappings. A distinguishing feature of FairerCLIP~\cite{dehdashtian2024fairerclip} is its ability to debias the representations without needing ground-truth labels $Y$ and $S$. Data rebalancing~\cite{alabdulmohsin2024clip} has also been explored as an alternative to model debiasing. Although these debiasing approaches have focused on zero-shot image classification and image-text retrieval tasks, understanding and mitigating biases in aligned image-text representations, have far greater implications, as CLIP-style models are commonly used as feature extractors in large multimodal models and text-to-image generation.

\underline{\emph{Large Multimodal Models:}} A few recent efforts~\cite{cui2023holistic,zhang2024debiasing} have also been made to uncover biases in large multimodal models (LMMs), such as GPT-4V~\cite{achiam2023gpt} and LLaVA~\cite{liu2023visual}, capable of more versatile tasks including captioning and VQA. Owing to the variety of tasks and the rapid evolution of multimodal architectures, work in this area is still scarce and does not fully capture the complexities of LMM fairness.

Apart from the initial forays discussed above, understanding and mitigating biases in foundation models largely remain an open problem. This state of affairs offers both new opportunities and new challenges to the computer vision community. First, foundation models are often applied to solve downstream tasks in \emph{zero-shot} or \emph{few-shot} settings, which have not been the subject of much study in the debiasing literature. Second, while this literature typically addresses the fairness of specific tasks, there are no unified measures of fairness for the diverse tasks and contexts on which foundation models are evaluated. Finally, the most widely adopted notions of fairness are defined with respect to \emph{closed vocabularies} of target labels and sensitive attributes. This is insufficient to quantify the fairness of foundation models, which target \emph{open} set applications using vocabularies based on natural language.

\section{Concluding Remarks}

In this paper, we reviewed the advancements made by the research community in measuring and mitigating bias w.r.t. sensitive protected groups in various computer vision tasks. Specifically, we discussed the social and technical origins of bias in computer vision systems, the different definitions of fairness considered by the research community, and different bias mitigation techniques and benchmark datasets to evaluate and compare them. Finally, we discussed fairness and bias in the context of modern multimodal foundation and generative models. We hope this survey provides a helpful and detailed overview for new researchers and practitioners, provides a convenient reference for relevant experts, and encourages future progress in the informed design of fair and equitable computer vision systems.

\vspace{0.05in}
\noindent{\bf Acknowledgments. } This work was funded by NSF Awards No. 2221943, 2041009, and 2147116 and gift funding from Amazon through the NSF Program on Fairness in Artificial Intelligence in Collaboration with Amazon. We also thank the organizers of the NSF-Amazon PI Meeting in Arlington, VA, on January 2024.

\bibliography{main}

\vspace{-3em}
\begin{IEEEbiography}[\vspace{-4em}{\rotatebox{-90}{\includegraphics[width=0.9in,height=1.1in,clip,keepaspectratio]{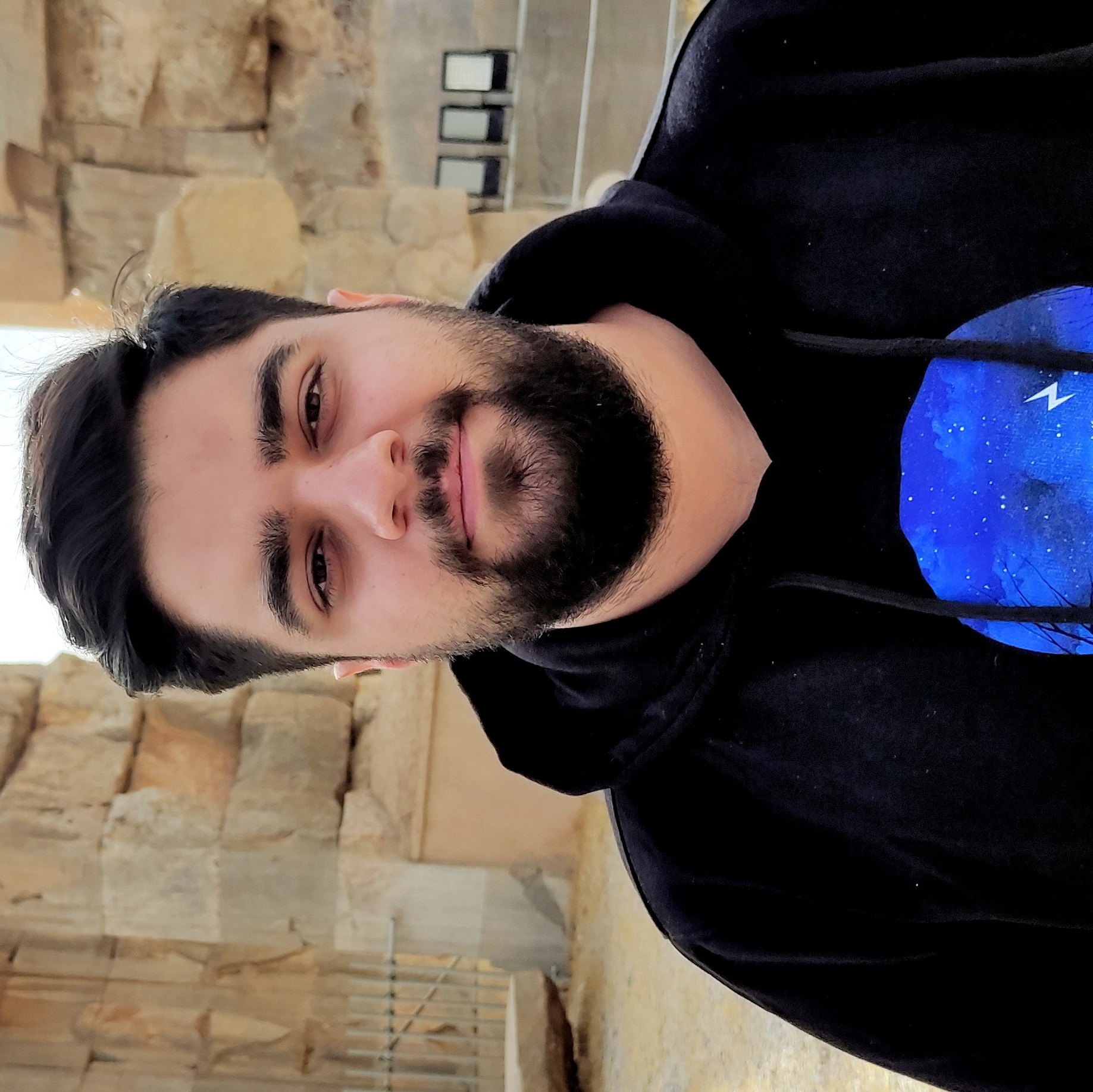}}}]{Sepehr Dehdashtian} is a Ph.D. student in the Department of Computer Science and Engineering at Michigan State University. He received a MSc degree in Electrical Engineering from Sharif University of Technology, Iran. His research interests are in Responsible AI and Fairness in Multimodal and Generative models.
\end{IEEEbiography}

\vspace{-6em}
\begin{IEEEbiography}[\vspace{-3.3em}{{\includegraphics[width=0.9in,height=1.1in,clip,keepaspectratio]{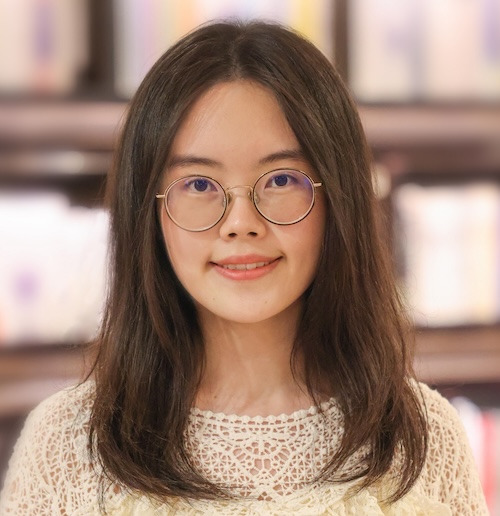}}}]{Ruozhen (Catherine) He} is a Ph.D. Student in the Department of Computer Science at Rice University. She received a BSc degree (First Class Honors) from the City University of Hong Kong. Her primary research interests lie in computer vision, focusing on efficient algorithms for multimodal models under limited supervision.
\end{IEEEbiography}

\vspace{-6em}
\begin{IEEEbiography}[\vspace{-3.3em}{{\includegraphics[width=0.9in,height=1.1in,clip,keepaspectratio]{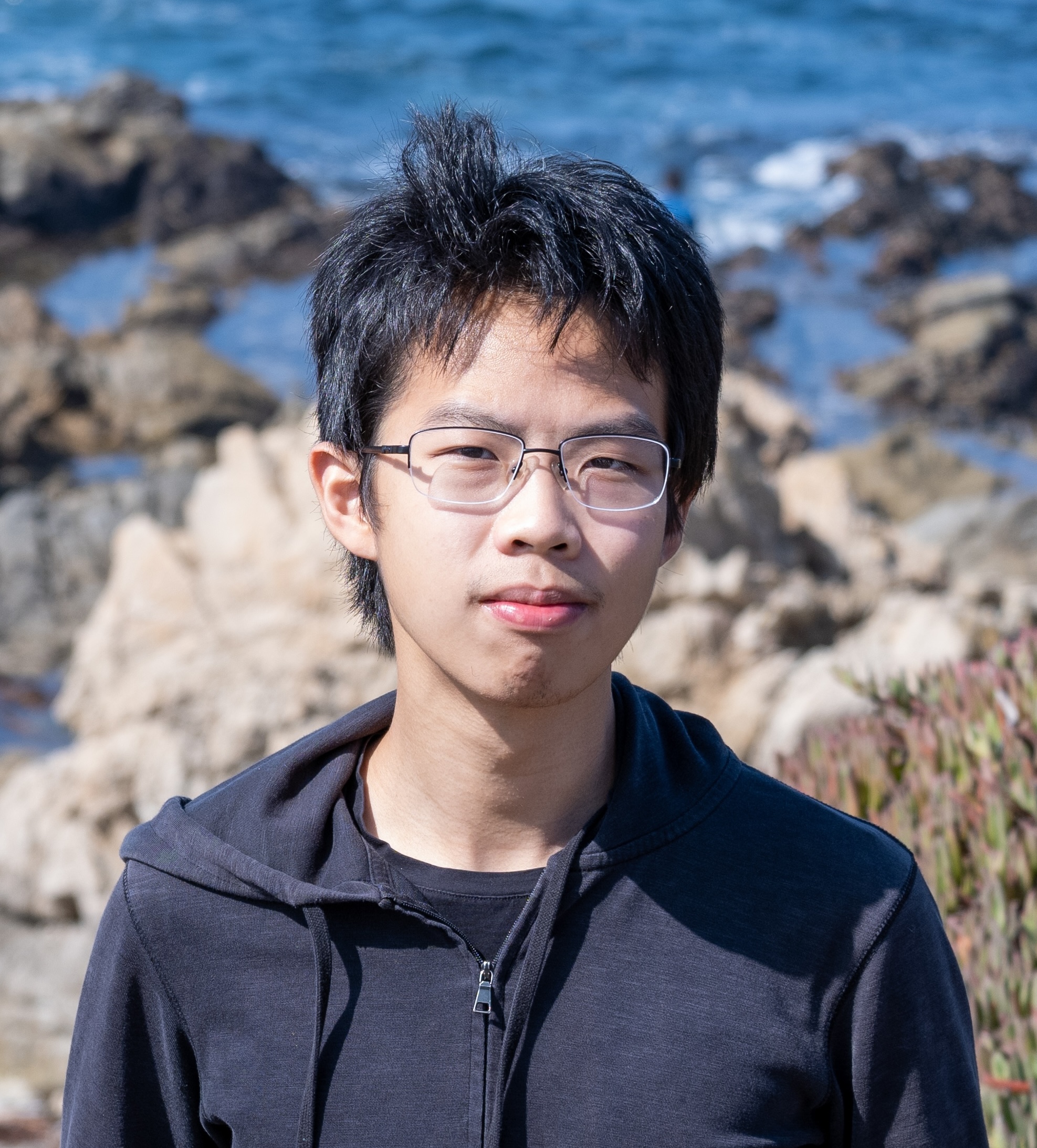}}}]{Yi Li} is a Ph.D. candidate in the Department of Electrical and Computer Engineering at the University of California San Diego. He received his B.Eng. degree (First Class Honors) from the Chinese University of Hong Kong. His research interests include representation learning and bias mitigation for computer vision and multimodal machine learning.
\end{IEEEbiography}

\vspace{-6em}
\begin{IEEEbiography}[\vspace{-1.2em}{{\includegraphics[width=0.9in,height=1.1in,clip,keepaspectratio]{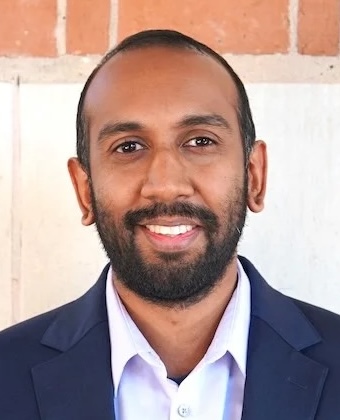}}}]{Guha Balakrishnan} is an Assistant Professor of Electrical and Computer Engineering working in the fields of computer vision and graphics. He is interested in the theory, practical design, and downstream applications of generative models for complex visual data. He is particularly excited by their application to promote fairness and accountability in vision systems. He received a Ph.D. in EECS from MIT in 2018.
\end{IEEEbiography}

\vspace{-5em}
\begin{IEEEbiography}[{\includegraphics[width=1in,height=1.25in,clip,keepaspectratio]{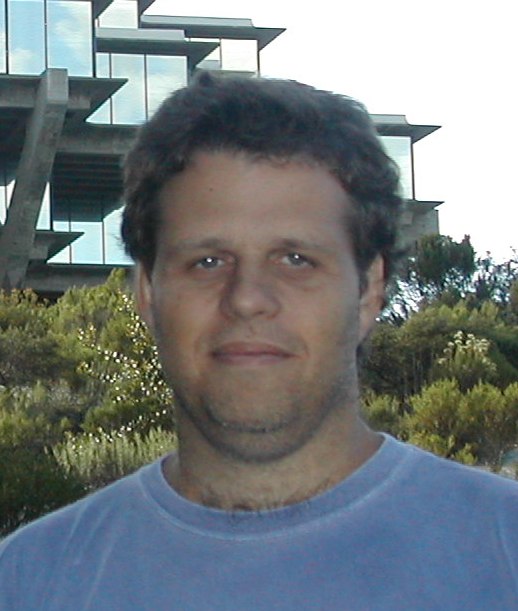}}]{Nuno Vasconcelos (Fellow, IEEE)} is a Professor of Electrical and Computer Engineering at the University of California, San Diego, where he heads the Statistical Visual Computing Laboratory. He has received an NSF CAREER award, a Hellman Fellowship, several best paper awards, and authored more than 200 peer-reviewed publications. He has been the Area Chair of multiple computer vision conferences and the Associate Editor of the IEEE Transactions on PAMI.
\end{IEEEbiography}

\vspace{-5em}
\begin{IEEEbiography}[\vspace{-1.2em}{{\includegraphics[width=0.9in,height=1.1in,clip,keepaspectratio]{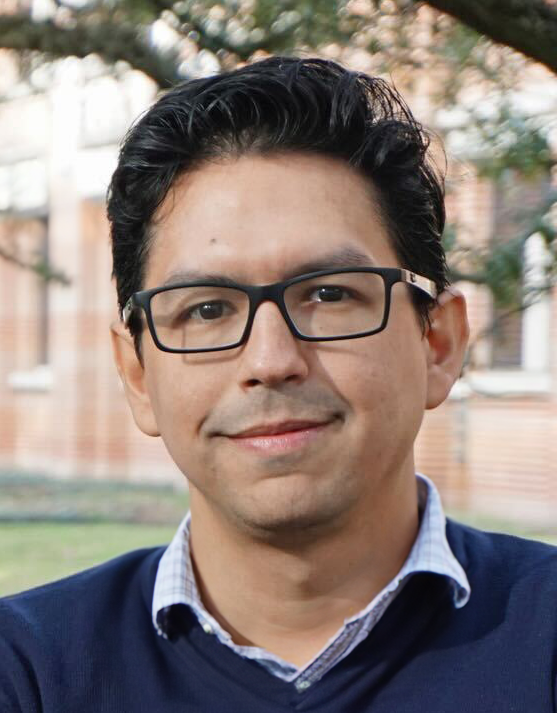}}}]{Vicente Ord\'o\~nez (Member, IEEE)} is an Associate Professor in the Department of Computer Science at Rice University. His research interests are at the intersection of Computer Vision and Natural Language Processing. He received a Ph.D. in Computer Science from the University of North Carolina at Chapel Hill in 2015. He received the Marr Prize at ICCV 2013 and a Best Paper award at EMNLP 2017. 
\end{IEEEbiography}

\vspace{-5em}
\begin{IEEEbiography}[{\includegraphics[width=0.9in,height=1.1in, clip,keepaspectratio]{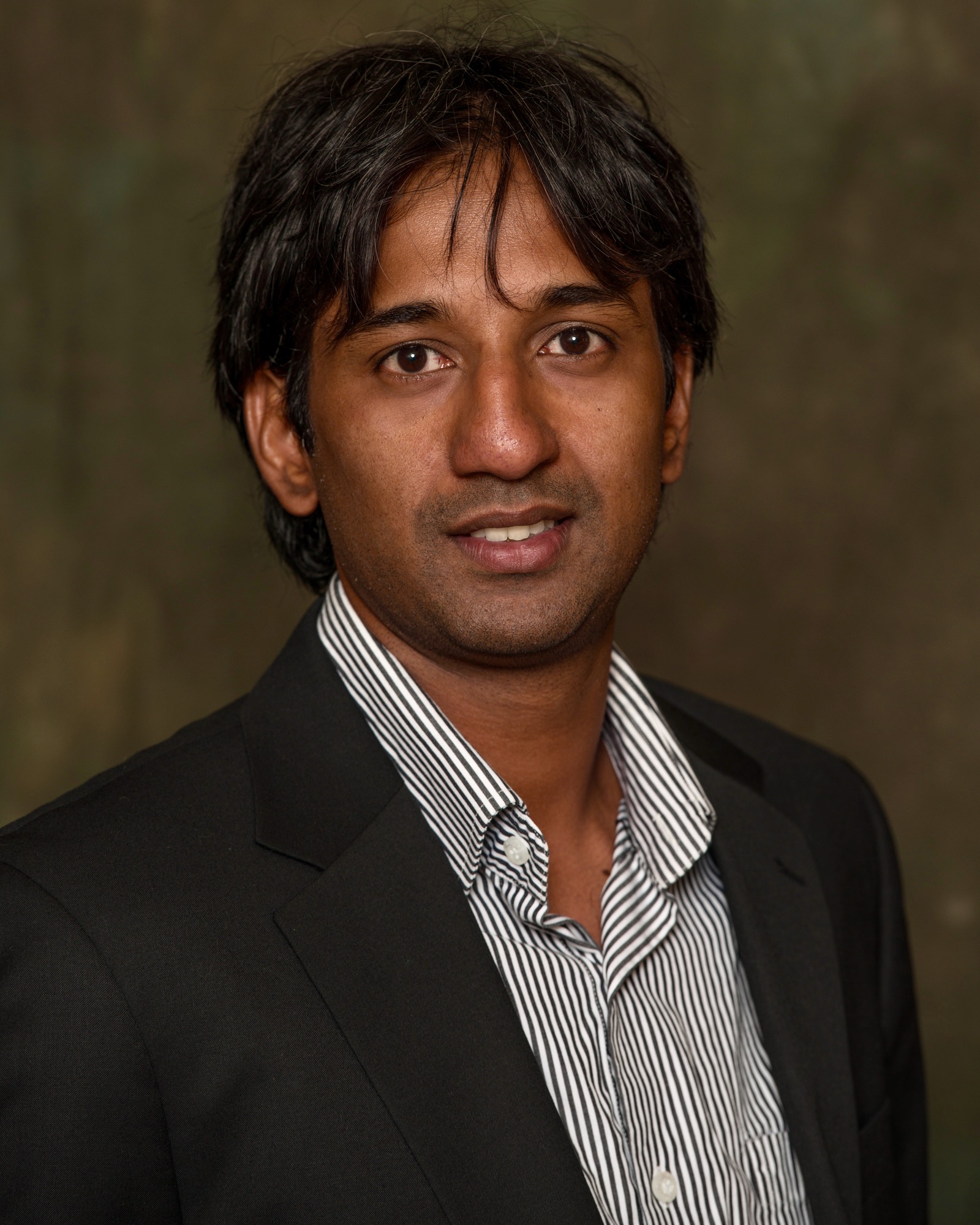}}]{Vishnu Naresh Boddeti (Member, IEEE)} is an Associate Professor in the Department of Computer Science and Engineering at Michigan State University. He received a Ph.D. in Electrical and Computer Engineering from Carnegie Mellon University. His research interests are Computer Vision, Pattern Recognition, and Machine Learning. Papers co-authored by him have received Best Paper Awards at BTAS 2013 and GECCO 2019 and Best Student Paper Awards at ACCV 2018, SMAIS 2022, IJCB 2022, and TBIOM 2023.
\end{IEEEbiography}

\end{document}